\documentclass[11pt]{article}

\usepackage[final]{acl}

\usepackage{times}
\usepackage{latexsym}

\usepackage[T1]{fontenc}
\usepackage[utf8]{inputenc}

\usepackage{microtype}

\usepackage{inconsolata}

\usepackage{graphicx}
\usepackage{booktabs}
\usepackage[table]{xcolor}
\usepackage{booktabs,multirow,array}
\usepackage[font=footnotesize,hangindent=10pt]{subfig}
\usepackage{subcaption}
\usepackage{amssymb}  % 提供 \triangleq 命令
\usepackage{amsmath}  % 提供数学环境
\usepackage[switch]{lineno}
\usepackage[most]{tcolorbox}
\usepackage{listings}
\tcbuselibrary{listings,breakable}
\usepackage{tabularx}
\usepackage{enumitem}
\usepackage{placeins}
\usepackage{fancyhdr}
\fancypagestyle{firstpageaccepted}{
  \fancyhf{}

  \fancyfoot[C]{%
    \parbox{\textwidth}{%
      \centering\footnotesize
      \textit{Proceedings of the 2026 Conference on Empirical Methods in Natural Language Processing.}%
    }%
  }
}
\definecolor{promptpurple}{RGB}{160,60,200}
\newtcolorbox{promptbox}{
  enhanced,
  colback=white,
  colframe=black,
  boxrule=0.6pt,
  sharp corners,
  left=2mm,right=2mm,top=2mm,bottom=2mm,
  boxsep=0mm,
}
\newcommand{\promptlabel}[1]{\textcolor{promptpurple}{\textbf{#1}}}

\lstdefinestyle{casejson}{
  basicstyle=\ttfamily\footnotesize,
  columns=fullflexible,
  breaklines=true,
  breakatwhitespace=true,
  keepspaces=true,
  showstringspaces=false,
  tabsize=2,
  moredelim=**[is][\color{red}]{@@}{@@},
}

\usepackage{circledsteps}
\usepackage{caption}
\newcommand{\circledtext}[1]{%
  \textcircled{\scriptsize #1}%
}

\usepackage[capitalise]{cleveref}
\crefformat{section}{\S#2#1#3}

\usepackage{tabularray}
\UseTblrLibrary{booktabs}

\newcommand{\bullethdr}[1]{\noindent\textbullet\,\textbf{#1}}

\newcommand{\header}[1]{\smallskip\noindent\textbf{#1}}

\newcommand{\indr}[1]{\mathbf{1}(#1)}
\newcommand{\CF}{\mathcal{F}}
\newcommand{\CD}{\mathcal{D}}
\newcommand{\CV}{\mathcal{V}}
\newcommand{\CE}{\mathcal{E}}
\newcommand{\CX}{\mathcal{X}}
\newcommand{\CG}{\mathcal{G}}
\newcommand{\CP}{\mathcal{P}}

\title{FloCA: Towards Faithful and Logically Consistent Flowchart Reasoning}

\author{
  Jinzi Zou$^{1}$ \quad
  Bolin Wang$^{2}$ \quad
  Shuo Zhang$^{1}$ \quad
  Nuo Xu$^{1}$ \quad
  Junzhou Zhao$^{1}$\thanks{Corresponding author.}
  \\
  $^{1}$MoE KLINNS Lab, Xi'an Jiaotong University, China
  \\
  $^{2}$China Mobile Group Shaanxi Co., Ltd.
  \\
  \texttt{\{jinzizou,zs412082986\}@stu.xjtu.edu.cn}
  \\
  \texttt{nxu@sei.xjtu.edu.cn},
  \texttt{junzhou.zhao@xjtu.edu.cn}
  \\
  \texttt{wangbolin2@sn.chinamobile.com}
}

\begin{document}
\maketitle
% 首页底部的接收声明（无脚注编号）
\thispagestyle{firstpageaccepted}
\begin{abstract}
Flowchart-oriented dialogue (FOD) systems aim to guide users through multi-turn
decision-making or operational procedures by following a domain-specific
flowchart to achieve a task goal.
In this work, we formalize flowchart reasoning in FOD as grounding user input to
flowchart nodes at each dialogue turn while ensuring node transition is
consistent with the correct flowchart path.
Despite recent advances of LLMs in task-oriented dialogue systems, adapting them
to FOD still faces two limitations: (1) LLMs lack an explicit mechanism to
represent and reason over flowchart topology, and (2) they are prone to
hallucinations, leading to unfaithful flowchart reasoning.
To address these limitations, we propose FloCA, a zero-shot flowchart-oriented
conversational agent.
FloCA uses an LLM for intent understanding and response generation, while
delegating flowchart reasoning to an external tool that performs
topology-constrained graph execution, ensuring faithful and logically consistent
node transitions across dialogue turns.
We further introduce an evaluation framework with an LLM-based user simulator
and five new metrics covering reasoning accuracy and interaction efficiency.
Extensive experiments on FLODIAL and PFDial datasets highlight the bottlenecks
of existing LLM/VLM-based methods and demonstrate the superiority of FloCA. The code and dataset are publicly available at \url{https://github.com/Jinzi-Zou/FloCA-flowchart-reasoning}.
\end{abstract}

\section{Introduction}

\begin{figure}[t]
  \centering
  \subfloat[LLM/VLM-based methods\label{fig:intro_example_a}]{%
    \includegraphics[width=.8\linewidth]{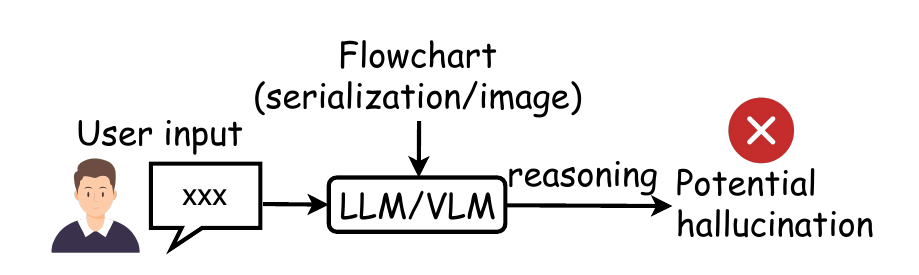}}\\
  \subfloat[FloCA\label{fig:intro_example_b}]{%
    \includegraphics[width=.8\linewidth]{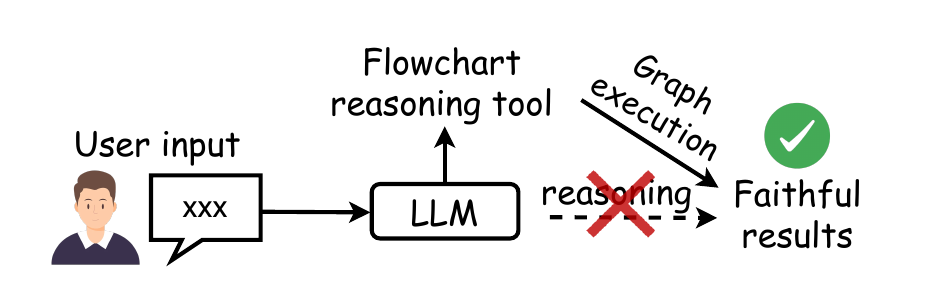}}
  \caption{Comparison of workflows for the existing methods and
      FloCA.
  (a) relies on the intrinsic reasoning ability of LLMs or VLMs, which suffers from hallucination, and (b) leverages a flowchart reasoning tool to
  preserve topological integrity and ensure faithful and logically consistent
  reasoning.}
  \label{fig:intro_example}
\end{figure}

Flowcharts are widely used to guide users through decision-making and procedural
steps, especially in domains such as
troubleshooting~\cite{raghu2021end,raghu2022structural,yamanaka2025flowchart},
medical diagnosis~\cite{li2023meddm,kim2024mdagents,xu2025using}, and legal
reasoning~\cite{mclachlan2021visualisation,onami-etal-2025-legalviz}.
Unlike textual guides, flowcharts often require users' strict adherence to their
prescribed path to reach the final goal.
For example, in troubleshooting, each node in the flowchart corresponds to a
specific diagnostic question that leads to different branches.
Making an incorrect judgment at any decision node can lead to an incorrect
diagnosis.
However, real-world flowcharts are often quite complex and domain-specific,
making it difficult for non-expert users to follow them correctly without
external guidance.

Raghu et al.~\shortcite{raghu2021end} propose FLONET, an end-to-end
flowchart-based dialogue system for troubleshooting.
However, FLONET lacks mechanisms to ensure the model follows the correct
decision path and thus may result in unfaithful reasoning.
This highlights the need for a task definition that explicitly requires faithful
and logically consistent reasoning.
To fill this gap, we formally define a new task of
\textbf{F}lowchart-\textbf{O}riented \textbf{D}ialogue (\textbf{FOD}),
specifying its core objectives, and introduce both a novel method and an
evaluation framework tailored to this task.

Despite recent advances in using Large Language Models (LLMs) for task-oriented
dialogue
(TOD)~\cite{li2024large,xu2025agenttod,baidya2025behavior,acikgoz2025desideratum},
applying them to FOD remains challenging, primarily due to two core
difficulties.
\textbf{First}, LLMs lack an explicit mechanism to represent and reason over
complex flowchart topology.
\textbf{Second}, LLMs are prone to struggle with complex reasoning, leading to
incorrect and logically inconsistent node transitions in multi-turn dialogues.

Prior work has attempted to mitigate these issues through three main strategies:
(1) \emph{Retrieval-Augmented Generation (RAG)-based
  methods}~\cite{raghu2022structural,soman2025graph}, which encode path
attributes as embeddings for retrieval; (2) \emph{Visual Language Models
  (VLMs)-based
  methods}~\cite{singh2024flowvqa,zhang2024first,he2025flow2code,suri2025follow},
which use VLMs to reason directly over flowchart images; (3) \emph{Graph
  serialization methods}~\cite{ye2024language,xu2025graphomni,yin2025talk},
which linearize flowcharts into textual sequences to input LLMs.
Among them, RAG-based methods are sensitive to retrieval accuracy; VLM-based
methods may fail to ground decision logic to specific visual elements; graph
serialization methods flatten the topology, forcing LLMs to reason over a lossy
representation.
Other efforts to enhance LLM reasoning through
prompting~\cite{liu2025logic,yao2023tree,besta2024graph},
fine-tuning~\cite{wan2024logicasker,morishita2024enhancing,bao2024abstract,zhang-etal-2025-pfdial},
or symbolic solvers~\cite{pan2023logic,ryu2025divide,callewaert2025verus} either
remain probabilistic, require costly data, or cannot handle interactive,
multi-turn tasks.
Moreover, a common thread persists, i.e., relying on the intrinsic capabilities
of LLMs or VLMs does not eliminate the underlying risk of hallucination in node
transitions.

To overcome the above limitations, we propose \textbf{FloCA}, a zero-shot
autonomous \textbf{Flo}wchart-oriented \textbf{C}onversational \textbf{A}gent.
FloCA leverages an LLM for user intent understanding, semantic matching, and
response generation, while delegating flowchart reasoning to a flowchart
reasoning tool.
This tool performs topology-constrained graph execution, ensuring faithful and
logically consistent node transitions throughout the dialogue
(\cref{fig:intro_example}).
To evaluate FloCA in realistic scenarios, we also introduce a novel evaluation
framework equipped with an LLM-powered user simulator and five new metrics that
assess both reasoning accuracy and interaction efficiency.
Extensive experiments on the FLODIAL and PFDial datasets show that FloCA
significantly outperforms existing LLM-based and VLM-based methods, establishing
a strong baseline for future FOD research.
Our contributions are summarized below:

\begin{itemize}\itemsep0pt
\item To the best of our knowledge, we are the first to formally define the task
  of \textbf{Flowchart-Oriented Dialogue}, where the main objective is to guide
  the user to make decisions and conduct operations step-by-step along the
  correct path within flowcharts (\cref{sec:problem}).
\item We design a new flowchart-oriented conversational agent \textbf{FloCA},
  which is equipped with a faithful reasoning tool to ensure logically
  consistent flowchart reasoning (\cref{sec:method}).
\item We propose a new evaluation framework for the FOD task, bridging the gap
  left by traditional TOD evaluation settings that are incompatible with the FOD
  task (\cref{sec:evaluation_framework}).
\item We conduct extensive experiments on FLODIAL and PFDial datasets.
  Empirical results show that FloCA achieves the highest task success rate among
  all LLM-based and VLM-based baselines, establishing it as a strong baseline
  for future research.
  (\cref{sec:experiments}).
\end{itemize}

\section{Related Work}

\header{LLMs for Flowchart Reasoning}.
Recent research about flowchart reasoning can be broadly divided into three
categories.
(1) \emph{Visual question answering}.
These methods input flowchart images into VLMs to perform logical reasoning
tasks.
Several benchmarks such as FlowchartQA~\cite{tannert2023flowchartqa},
FlowCE~\cite{zhang2024first}, FlowLearn~\cite{pan2024flowlearn},
SCI-CQA~\cite{shen2024rethinking} and FlowVQA \cite{singh2024flowvqa}, evaluate
VLMs' capabilities in visual logic and spatial reasoning.
TEXTFLOW~\cite{ye2025beyond} converts flowchart images into text via VLMs, then
performs reasoning in the language space.
FlowPathAgent~\cite{suri2025follow} further enhances interpretability by mapping
attribution paths from the answer back to relevant flowchart regions.
(2) \emph{Flowchart recognition and code generation}.
These methods focus on recognizing flowchart images from user
prompts and translating them into executable code in various programming
languages~\cite{herrera2017flow2code,liu2022code,shukla2023towards,mentari2024automatic,he2025flow2code}.
(3) \emph{Decision-making}.
Flowcharts have been leveraged to guide LLMs in multi-step decision-making across diverse domains, including clinical diagnosis \cite{li2023meddm}, reasoning enhancement via visual context \cite{wang2024novel}, extraction of decision flows from dialogues \cite{yamanaka2025flowchart}, and code repair for nullability errors \cite{karimipour2025llm}.
Despite recent advances, existing approaches primarily treat flowcharts as
enhanced external knowledge or prompting context for LLMs, which still rely on
probabilistic generation and remain vulnerable to hallucination.
In contrast, our work treats the flowchart as a structured reasoning tool that
explicitly constrains the agent's reasoning trajectory along the flowchart path,
effectively mitigating hallucination.

\header{Autonomous LLM-based Conversational Agents}.
Autonomous LLM-based agents decompose complex tasks into manageable subtasks by
planning, making decisions, and invoking external tools or APIs when needed.
Recent works such as ReAct~\cite{yao2022react},
Reflexion~\cite{shinn2023reflexion}, Toolformer~\cite{schick2023toolformer},
AutoGPT~\cite{RichardsAutogpt2023}, BabyAGI~\cite{NakajimaBabyagi2023},
AgentLM~\cite{zeng2024agenttuning}, Microsoft's Jarvis
(HuggingGPT)~\cite{shen2023hugginggpt}, and AgentGPT~\cite{reworkdAgentGPT2025},
explore various architectures for enabling such capabilities.
More recently, conversational agents that integrate dialogue-based planning,
goal tracking, and adaptive decision-making have shown promising results in
domains such as healthcare~\cite{yang2024talk2care,abbasian2025conversational},
medical diagnosis~\cite{tu2025towards,chen2025enhancing},
recommendation~\cite{huang2025recommender}, and TOD
systems~\cite{xu2024rethinking,xu2025agenttod,baidya2025behavior}.
However, existing work has largely overlooked the flowchart-oriented dialogue
task.
To fill this gap, we propose a faithful conversational agent that supports
topology-constrained flowchart reasoning, offering a new conversational agent
design paradigm for scenarios that demand strict adherence to procedural logic.

\begin{figure*}[t]
  \centering
  \includegraphics[width=.9\linewidth]{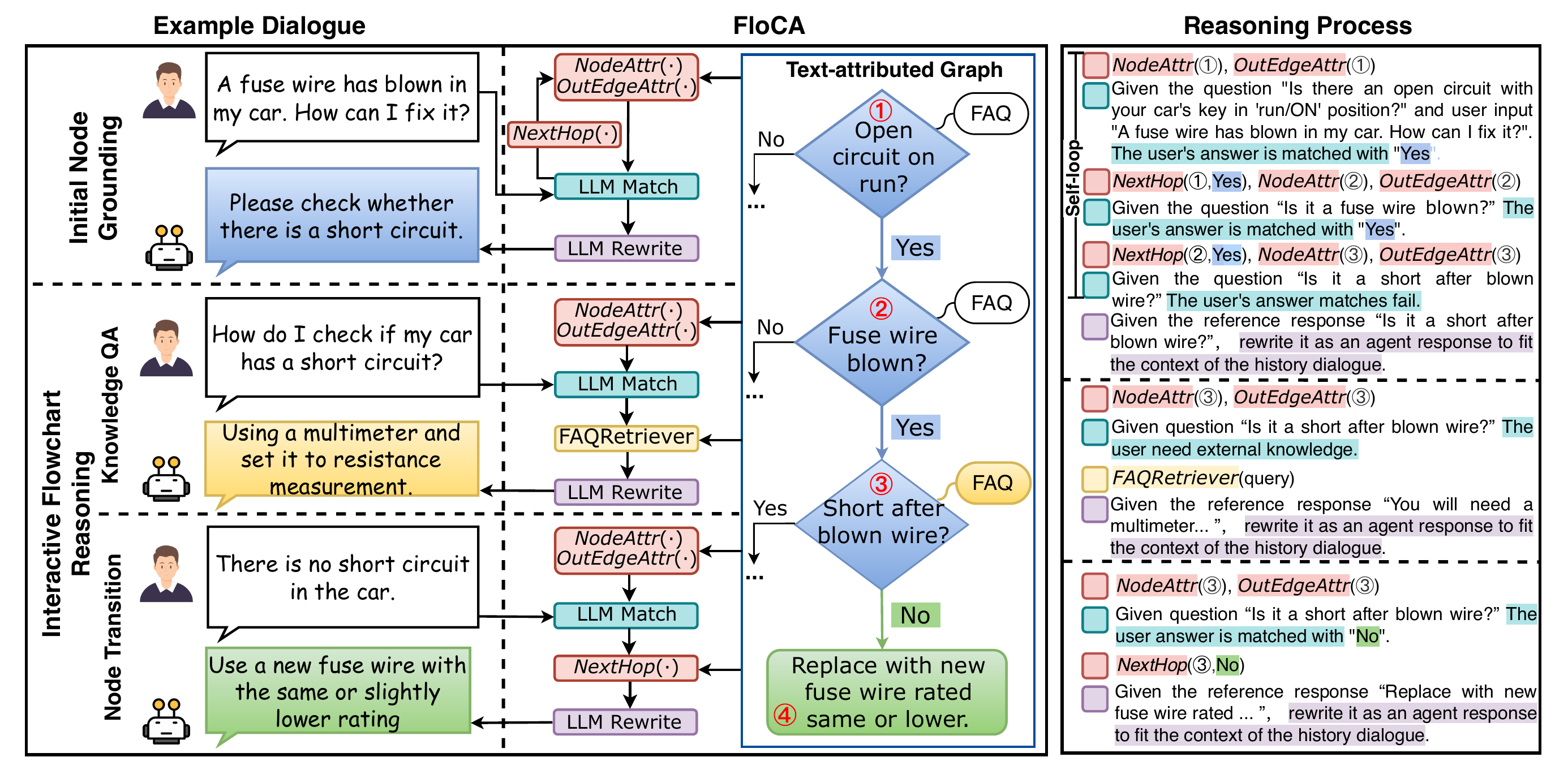}
  \caption{An overview of FloCA. FloCA consists of two core components: an
    instruction-following LLM and a faithful flowchart reasoning tool.
    The left figure shows an example dialogue in troubleshooting, where each
    agent output corresponds to the result of the flowchart reasoning for a
    specific flowchart node or an answer from FAQ database.
    The right figure depicts the entire multi-turn reasoning process, with
    colors representing the graph functions and LLM reasoning processes, showing the
    input and output at each step and how reasoning is carried out.
\label{fig:overview}}
\end{figure*}

\section{Flowchart-oriented Dialogues}
\label{sec:problem}

We formalize the flowchart-oriented dialogue (FOD) task as follows.
A flowchart is represented as a graph $\CF=(\CV, \CE, \CX)$, where $\CV=\{v_1,
\ldots, v_{|\CV|}\}$ is a set of nodes, $\CE=\{e_1, \ldots, e_{|\CE|}\}$ is a
set of edges, and $\CX=\{x_v\mid v\in\CV\}\cup\{x_e\mid e\in\CE\}$ denotes the
textual attributes attached to nodes and edges.
Each edge $e=(v_i,v_j)\in\CE$ specifies a transition from a source node $v_i$ to
a target node $v_j$, and the associated edge attribute encodes the condition
required to move from $v_i$ to $v_j$.
This transition corresponds to a decision-making step or the execution of
specific actions and procedures required to progress along the flowchart.
For example, in the troubleshooting flowchart shown in \cref{fig:overview}, each
node represents a diagnostic question, such as ``Fuse wire blown?''
and ``Short after blown wire?'', associated with some text describing the
trouble in detail.
An edge connecting two nodes represents a possible transition during
troubleshooting, e.g., if the user reports ``Fuse wire has blown'', then the
agent may recommend the user to check whether there is a short circuit.

A flowchart-oriented dialogue $\CD$ consists of $T$ turns of alternating user
and agent utterances, denoted by $\CD=\{(c_1^u,c_1^a), \ldots, (c_T^u,c_T^a)\}$,
where $(c_t^u,c_t^a)$ denotes the user and agent utterances at the $t$-th turn.
We assume the agent acts after observing the user utterance $c_t^u$.
Given a flowchart $\CF$ and the dialogue history up to the $t$-th turn,
$\CD_t=\{(c_1^u,c_1^a), \ldots, (c_{t-1}^u,c_{t-1}^a), c_t^u\}$, where $1\le t
\le T$, the agent pursues the following two objectives.

\bullethdr{Flowchart reasoning}.
The agent should first ground the current dialogue context to a node
$\hat{v}_t\in\CV$ in the flowchart.
The inferred node sequence across turns $(\hat{v}_1,\hat{v}_2,\ldots,\hat{v}_t)$
must remain consistent with the flowchart topology, i.e., either
$(\hat{v}_k,\hat{v}_{k+1})\in\CE$ or $\hat{v}_k=\hat{v}_{k+1}$ for all
$k=1,\ldots,t-1$.
For example, if $\hat{v}_{k}$ is node ``Open circuit on run'', then the
$\hat{v}_{k+1}$ can be ``Fuse wire blown check'' while ``Short after blown
wire'' cannot.

% $(\hat{v}_k,\hat{v}_{k+1})\in\CE$ for all $k=1,\ldots,t-1$.
% The inferred node sequence $(\hat{v}_1,\hat{v}_2,\ldots,\hat{v}_t)$ during
% multi-turn dialogue must be consistent with the flowchart transitions, i.e.,
% $(\hat{v}_k,\hat{v}_{k+1})\in\CE$ for all $k=1,\ldots,t-1$.

\bullethdr{Response generation}.
Conditioned on $\mathcal{D}_t$ and the grounded node $\hat{v}_t$, the agent
generates a context-aware utterance $\hat{c}_t^a$ based on the attributes
$x_{\hat{v}_t}$.
For example, if the grounded node is ``Fuse wire blown?'', then the agent should
follow the flowchart and generate a response about checking whether the fuse
wire has blown.

\section{FloCA: A Flowchart Reasoning Agent}
\label{sec:method}

We propose FloCA, a zero-shot LLM-based autonomous conversational agent for the FOD task.
\cref{fig:overview} shows the workflow of FloCA with a troubleshooting example.
FloCA consists of an instruction-following LLM and a faithful flowchart reasoning tool.
The key idea of FloCA is to delegate topology-constrained flowchart reasoning to the tool, while the LLM handles user intent understanding, semantic matching, and response generation.
During interaction, FloCA addresses three key challenges: initial node grounding, interactive flowchart reasoning, and domain knowledge QA.
We first introduce the flowchart reasoning tool and then describe how FloCA addresses each challenge.

\subsection{Flowchart Reasoning Tool}

To enable faithful flowchart reasoning and ensure correct node grounding at each
dialogue turn, we introduce a flowchart reasoning tool that guarantees
deterministic transitions according to the flowchart topology.
To preserve both the topological structure and textual information of the
flowchart, we model the tool as a text-attributed graph with four key graph
functions:

\begin{itemize}\itemsep0pt
\item $\texttt{NodeAttr}(\mathsf{NodeID})$: Returns the textual attribute of a
  given node.
\item $\texttt{OutEdgeAttr}(\mathsf{NodeID})$: Returns a list of textual
  attributes for all outgoing edges of the given node.
\item $\texttt{NextHop}(\mathsf{NodeID}, \mathsf{EdgeAttr})$: Given a source
  node and the textual attribute of one outgoing edge, returns the connected
  target node ID.
\item $\texttt{TerminalCheck}(\mathsf{NodeID})$: Returns $1$ if the node is
  terminal and $0$ otherwise.
\end{itemize}

FloCA uses the LLM to interpret user utterances, analyze intent, and determine
which transition condition is supported by the user's input, while delegating
node transitions to the reasoning tool.
Given the currently grounded node, FloCA uses the LLM to rewrite the node
attribute into a natural language response, which may either present a decision
question or explain a required operation.
We next describe how FloCA combines the LLM and the reasoning tool to handle
three key dialogue challenges.

\subsection{Initial Node Grounding}

At the beginning of a dialogue, the user's first utterance often conveys their request or current situation.
In some cases, this utterance already provides
information that implicitly resolves early decision questions, allowing the
dialogue to be grounded at an intermediate node rather than always starting from the root.
For example, as shown in \cref{fig:overview}, if a user reports that a fuse wire has already blown, the agent can begin reasoning from the node \circledtext{3} rather than node \circledtext{1}.
To avoid redundant interaction turns,
we introduce a self-loop mechanism for initial node grounding.
Internally, FloCA starts the self-loop from the root and calls the $\texttt{NodeAttr}$ and
$\texttt{OutEdgeAttr}$ functions to retrieve the attributes of the current node and its outgoing edges.
For a decision node, the LLM matches the user's utterance against the
transition conditions of the outgoing edges.
For an operation node, the LLM checks whether the user has already
completed the required action.
If a valid transition is identified, FloCA calls $\texttt{NextHop}$ to move to the
next node and iteratively continues the grounding process.
Otherwise, FloCA remains at the current node and uses
the LLM to rewrite the node attribute into a contextually appropriate response.

This mechanism allows FloCA to make effective use of the information in the
user's initial input, thereby reducing unnecessary interactions while ensuring
accurate initial grounding.

\subsection{Interactive Flowchart Reasoning}
\label{ss:reasoning}

After the initial node is grounded, FloCA continues the dialogue by
recursively applying the same graph execution logic at each subsequent turn.
Given the currently grounded node, FloCA calls $\texttt{NodeAttr}$ and
$\texttt{OutEdgeAttr}$ to retrieve the attributes of the node and its outgoing edges and uses the LLM to determine how the dialogue should proceed.
This interaction leads to three possible outcomes. (1) \emph{First}, if the user's utterance supports a valid outgoing transition, FloCA
calls $\texttt{NextHop}$ to move to the next node and continues the dialogue
from there. (2) \emph{Second}, if no valid transition is supported, FloCA remains at the current node and further clarifies the current decision question or required operation. (3) \emph{Third}, the user may be unable to answer the current decision question because
resolving it requires domain-specific knowledge beyond the user's expertise.
For example, as shown in \cref{fig:overview}, if the agent asks the user to check for a short circuit in a car,
the user may not know how to perform the check and instead ask a question such
as ``How do I check whether my car has a short circuit?''
In such cases, FloCA uses the LLM to identify the user's utterance as a
domain-specific knowledge query, suspends node transition, and calls a
retriever to fetch the relevant answer from an external FAQ database.
During this process, the currently grounded node remains unchanged until the
user provides the information required to determine a valid next-hop
transition.
FloCA then resumes flowchart reasoning from that node.

The above interaction continues until $\texttt{TerminalCheck}$ returns $1$,
indicating that a terminal node has been reached.

\section{Evaluation Framework for FOD Task}
\label{sec:evaluation_framework}

Prior works~\cite{raghu2021end,raghu2022structural} primarily evaluate flowchart TOD systems by comparing semantic similarity between generated responses and annotated dialogues. However, these methods overlook the agent's ability to follow the correct flowchart path.
As we explained in \cref{sec:problem}, faithful and logically consistent
flowchart reasoning is critical in addressing the FOD task.
To overcome these limitations, we propose a new evaluation framework that
incorporates an instruction-following LLM-based user simulator to support an
interactive reasoning environment.
Meanwhile, we also introduce five novel metrics to measure both reasoning accuracy
and interactive efficiency.

\subsection{User Simulator}

Inspired by prior work~\cite{xu2024rethinking}, we construct the user simulator based on an instruction-following LLM and design instruction prompts.
For each annotated dialogue sample, we extract user-known facts that are established at the start of the conversation from the dialogue, including observable situations and completed operations, and use them as scenario context in the prompt. 
These facts specify only the user's observable situation.
To further promote natural user language and realistic questioning behavior, we also include an annotated reference dialogue aligned with the scenario context.
The full prompt template and the process for extracting user-known facts are provided in Appendix \ref{App:user_simulator}.

\subsection{Flowchart Reasoning Metrics}

We evaluate the agent's ability in flowchart reasoning from three key aspects,
i.e., \emph{initial node reasoning}, \emph{terminal node reasoning}, and
\emph{path coverage}, which together assess both the accuracy and success of the
agent's flowchart reasoning.
Let $N$ denote the total number of test dialogue samples.
For the $i$-th sample, let $\CP_i=(\hat{v}_{i,1}, \ldots,\hat{v}_{i,\hat{T}_i})$ and $\CG_i=(v_{i,1}, \ldots, v_{i,T_i})$ be the predicted and ground-truth node sequences, respectively, where $\hat{T}_i$ and $T_i$ denote the number of interaction turns in the predicted and ground-truth dialogue.

\header{Initialization Metrics} measure how accurate the agent can identify the
initial node in the flowchart since each dialogue may not always start from the
root node.
Specifically, we introduce initial node grounding accuracy.

\bullethdr{Initial Node Grounding Accuracy (INGA)} measures whether the agent
identifies the correct initial node given the user's first utterance, i.e.,
\begin{equation}
  \mathrm{INGA}\triangleq\frac{1}{N}\sum_{i=1}^N\indr{\hat{v}_{i,1}=v_{i,1}},
\end{equation}
where $\indr{\cdot}$ is the indicator function.
Higher INGA suggests more reliable flowchart reasoning initialization.
This metric is crucial for decision-making flowcharts, since an incorrect initial node prediction can route the dialogue to a wrong branch and lead to task failure.

\header{Task Success Metrics} assess the accuracy of the predicted node sequences against the ground truth.
Specifically, we introduce two task success metrics.

\bullethdr{Terminal Node Grounding Accuracy (TNGA)} measures whether the agent
reaches the correct terminal node, i.e.,
\begin{equation}
  \mathrm{TNGA}\triangleq\frac{1}{N}\sum_{i=1}^N\indr{\hat{v}_{i,\hat{T}_i}=v_{i,T_i}}.
\end{equation}
A terminal node represents either the final solution or the last required step
of an operational procedure.
Higher TNGA indicates a higher likelihood of successful task completion.

\bullethdr{Path Coverage Accuracy (PCA)} measures whether the predicted node
sequence covers all ground truth nodes in order.
Note that only predicting the correct terminal node is insufficient to reflect
task success.
For example, in a passport application process, which involves legally mandated steps, PCA is essential because skipping an intermediate requirement, such as submitting a photo or providing a signature, invalidates the entire application.
Formally, we define PCA by
\begin{equation}
  \mathrm{PCA}\triangleq\frac{1}{N}\sum_{i=1}^N\indr{\CG_i\sqsubseteq\CP_i},
\end{equation}
where $\CG_i \sqsubseteq \CP_i$ indicates that $\CG_i$
is a subsequence of $\CP_i$.
Higher PCA demonstrating better logical consistency in flowchart reasoning.

\header{Efficiency Metrics} measures the flowchart interaction efficiency.
Specifically, we introduce two efficiency metrics.

\bullethdr{Node Stay Redundancy (NSR)} measures how often the agent stays at the
same flowchart node due to repeated clarifications within one-hop transition.
We define NSR by

\begin{equation}
  \mathrm{NSR} \triangleq \frac{1}{N} \sum_{i=1}^N \frac{\hat{T}_i - K_i}{\hat{T}_i},
\end{equation}
where $K_i$ is the number of predicted node transitions in the $i$-th sample, and $\hat{T}_i - K_i$ counts the redundant turns (assuming each transition ideally takes one turn). If $K_i = 0$ (no transition occurs), the term equals $1$.
Lower NSR indicates fewer redundant turns and more efficient interaction.

\bullethdr{Timeout Rate (TR)} measures the proportion of test samples that fail to
reach the terminal node within a turn budget $T_\tau$.
We define TR by
\begin{equation}
  \mathrm{TR}\triangleq\frac{1}{N}\sum_{i=1}^N\indr{\hat{T}_i>T_\tau}.
\end{equation}
Lower TR indicates that the agent can more reliably advance the dialogue to a
terminal node within the turn budget.

\section{Experiments}
\label{sec:experiments}

\subsection{Settings}
\begin{table*}[!t]
  \centering
  \footnotesize
  \begin{tblr}{
    width=.95\linewidth, rowsep=1pt, stretch=0,
    colspec={X[c]|X[2,l]|X[c]X[1.2,c]X[c]|X[c]X[c]|X[.5,c]X[.5,c]},
    cell{1}{1}={r=3,c=2}{},
    cell{1}{6,8}={c=2}{},
    cell{1,2}{3}={c=3}{},
    cell{2}{6-9}={r=2}{},
    cell{4}{1}={c=2}{},
    cell{5,11,21}{1}={r=6}{},
    cell{17}{1}={r=4}{},
    row{1-3}={font=\bfseries},
    column{1}={font=\bfseries},
    }
    \toprule
    Method
    &
    & Initialization Metrics
    &
    &
    & Task Success Metrics
    &
    & Efficiency Metrics
    & \\

    \cmidrule{3-9}
    &
    & INGA$\uparrow$
    &
    &
    & TNGA$\uparrow$
    & PCA$\uparrow$
    & TR$\downarrow$
    & NSR$\downarrow$ \\

    \cmidrule{3-5}
    &
    & Root-init$\uparrow$
    & Middle-init$\uparrow$
    & Overall$\uparrow$
    & & & & \\

    \midrule
    SA-FloNet~\cite{raghu2022structural} &
    & 93.71 & 41.79 & 67.75 & 1.65 & 0.20 & \textbf{0.00} & 23.24 \\
    \midrule

    Graph Serialization Methods & Llama-3.3-70B
    & \textbf{100.00} & 0.00 & 50.00 & 0.20 & 0.00 & 86.12  & 69.54 \\

    & Claude Opus 4.1
    & 98.57 & 23.13 & 60.85 & 83.43 & 80.12 & \textbf{0.00} & 0.38 \\

    & GPT-4o
    & 98.28 & 7.46 & 52.87 & 78.26 & 73.49 & \textbf{0.00} & 0.85 \\

    & GPT-5
    & 94.85 & \underline{67.91} & \textbf{81.38} & 87.16 & 79.29 & 0.62 & 1.15 \\

    & Qwen3
    & 78.00 & 2.23 & 40.11 & 67.70 & 54.24 & 0.41 & 4.34 \\

    & DeepSeek-R1
    & 89.71 & 15.67 & 52.69 & 76.81 & 66.66 & \textbf{0.00} & 1.39 \\
    \midrule

    % % \rowcolor{gray!15}
    RAG-enhanced Graph Serialization Methods
    & Llama-3.3-70B
    & 86.00 & 18.65 & 52.32 & 0.00 & 0.00 & 82.81 & 55.11 \\

    & Claude Opus 4.1
    & \underline{98.85} & 22.38 & 60.62 & 87.16 & 6.62 & \textbf{0.00} & 42.25 \\

    & GPT-4o
    & \underline{98.85} & 3.73 & 51.29 & 80.53 & 75.98 & \textbf{0.00} & 0.72 \\

    & GPT-5
    & 94.85 & \underline{67.91} & \textbf{81.38} & 87.16 & 79.29 & 0.62 & 1.15 \\

    & Qwen3
    & 88.57 & 9.70 & 49.13 & 71.84 & 51.12 & 4.14 & 5.94 \\

    & DeepSeek-R1
    & 93.71 & 14.92 & 54.31 & 76.81 & 66.66 & 1.03 & 1.10 \\
    \midrule

    Visual Language Models
    & Claude Opus 4.1
    & 85.71 & 12.68 & 49.20 & 16.35 & 3.72 & \underline{0.20} & 1.80 \\

    & Qwen-VL-Max
    & 71.14 & 5.97 & 38.55 & 36.23 & 12.62 & 4.34  & 5.18 \\

    & GPT-4o
    & 59.71 & 50.00 & 54.85 & 33.12 & 12.83 & 0.62  & 2.60 \\

    & GPT-5
    & 82.57 & \textbf{73.13} & \underline{77.85} & 50.10 & 29.39 & 9.10  & 8.45 \\
    \midrule

    FloCA
    & Llama-3.3-70B
    & 97.42 & 26.11 & 61.77 & 83.85 & 83.01 & \textbf{0.00} & 0.86 \\

    & Claude Opus 4.1
    & 97.42 & 24.62 & 61.02 & 91.30 & \underline{90.46} & \textbf{0.00} & \underline{0.23} \\

    & GPT-4o
    & \underline{98.85} & 25.37 & 62.11 & \underline{91.71} & \textbf{90.88} & \textbf{0.00} & \textbf{0.15} \\

    & GPT-5
    & 98.28 & 24.62 & 61.45 & \textbf{91.92} & \textbf{90.88} & \textbf{0.00} & 0.60 \\

    & Qwen3
    & 97.42 & 22.38 & 59.90 & \underline{91.71} & \textbf{90.88} & \textbf{0.00} & 0.64 \\

    & DeepSeek-R1
    & \textbf{100.00} & 20.89 & 60.44 & 90.26 & 89.84 & \textbf{0.00} & 0.45 \\
    \bottomrule
  \end{tblr}
  \caption{\emph{In-domain} FOD results on FLODIAL. All values are multiplied by
    100 for better readability. ``Root-init'' and ``Middle-init'' represent the
    subsets where the ground-truth initial node is the root node (350 samples) and a
    non-root node (134 samples) in a flowchart, respectively, and ``Overall'' is
    computed on all test samples. ``Task Success Metrics'' serve as the primary
    indicators of overall model performance. The {\bf bold font} highlights the best
    results, and the \underline{underline} highlights the second-best results.}
  \label{tab:results_on_FLODIAL}
\end{table*}

\header{Datasets}.
We use two public datasets, i.e., FLODIAL~\cite{raghu2021end} and
PFDial~\cite{zhang-etal-2025-pfdial}.
FLODIAL contains $12$ troubleshooting flowcharts and $2,738$ in-domain dialogues
split into $1,798/456/484$ for training/validation/testing.
PFDial contains $440$ training flowcharts, and $80$ flowcharts each for in-domain and out-of-domain testing in PlantUML format, from which we construct 513
in-domain and 416 out-of-domain synthetic dialogues based on the extracted flowchart paths.
Detailed dataset statistics and construction processes are provided in
Appendix~\ref{App:datastes}.

\header{Base Models}.
We implement FloCA based on a diverse set of language models, including GPT-4o, GPT-5, Claude Opus 4.1, DeepSeek-R1, Qwen3, Llama-3.3-70B, Llama-3.1-8B-Instruct, and Qwen2.5-7B-Instruct.

\header{Baselines}.
On FLODIAL, we compare FloCA with representative RAG-based,
graph-serialization-based, RAG-enhanced graph-serialization-based, and
VLM-based methods.
Specifically, these include SA-FloNet~\cite{raghu2022structural}, graph
serialization methods that directly provide serialized flowcharts to the model,
RAG-enhanced graph serialization methods built on the retriever in SA-FloNet,
and VLM-based methods that take flowchart images as input, including
Qwen-VL-Max, GPT-4o, GPT-5, and Claude Opus 4.1.
On PFDial, following its instruction-tuning setting ~\cite{zhang-etal-2025-pfdial}, we fine-tune Llama-3.1-8B and Qwen2.5-7B as baselines.

\header{Evaluation Tasks}.
We use the metrics in \Cref{sec:evaluation_framework} to evaluate the performance of FOD tasks in two settings, i.e., (1) in-domain, where the test flowcharts are seen during training; (2) out-of-domain, where the test ﬂowcharts are unseen during training.

\header{Implementation Details}.
For fair comparison, we use DeepSeek-R1 to construct the user simulator across
all evaluations and set the dialogue turn budget $T_\tau$ to twice the number of
ground-truth turns for each test sample.
We train SA-FloNet and fine-tune baselines on PFDial using the official hyperparameter settings from the original papers.
For FLODIAL, all methods use the same SA-FloNet retriever to handle
domain-knowledge QA so that performance differences mainly reflect flowchart reasoning ability rather than retrieval variation.

\subsection{Results on FLODIAL Dataset}

\Cref{tab:results_on_FLODIAL} reports the results of the FOD task on FLODIAL.
We present the main observations below.

\header{O1: FloCA achieves the highest overall task success}.
Although VLMs and graph-serialization methods obtain higher INGA, they perform much worse on TNGA and PCA. This is because taking the entire flowchart as input may help global semantic matching for initial node grounding, but subsequent reasoning still depends on probabilistic generation, requiring the model to implicitly maintain flowchart topology in language space. This often leads to semantically plausible but topologically invalid transitions, reducing task success. In contrast, FloCA uses a self-loop mechanism to ground the initial node more conservatively. When evidence is insufficient, it may remain at a node preceding the ground-truth initial node, which can slightly lower INGA but avoids skipping required steps and thus improves PCA. Moreover, FloCA performs reasoning through topology-constrained graph execution, explicitly preserving flowchart topology and yielding the most reliable end-to-end success.

\header{O2: Decoupling complex reasoning from LLMs reduces repeated
  clarification}.
FloCA achieves $\mathrm{TR}=0$ and the lowest NSR, indicating that it completes reasoning within the dialogue budget without repetitive clarification. Most baselines show much higher NSR, and RAG-based enhancement can even worsen repetition for strong LLMs such as Claude Opus 4.1, suggesting that misaligned retrieval may interfere with reasoning. FloCA avoids this by decoupling semantic understanding from flowchart reasoning, where LLM handles intent understanding, edge matching, and response generation, while the reasoning tool performs topology-constrained transitions. This separation improves robustness and supports stable, logically consistent progression across dialogue turns.

\begin{figure}[t]
  \centering
  \includegraphics[width=0.85\linewidth]{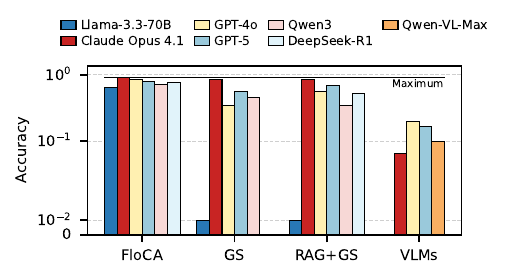}
  \caption{Flowchart reasoning accuracy around domain knowledge QA on
    FLODIAL. ``GS'', ``RAG+GS'' and ``VLMs'' denote baselines of graph serialization
    methods, RAG-enhanced graph serialization methods, and visual language models.}
  \label{fig:faq_evaluation}
\end{figure}

\header{Flowchart Reasoning Accuracy around Domain Knowledge QA}.
We evaluate whether an agent can correctly resume
flowchart reasoning after domain knowledge QA interruptions. Since domain
knowledge QA in FOD is intended to resolve user confusion and support dialogue
progression, the key ability is whether the agent can return to the pre-QA
flowchart state and make the correct next-hop transition after a single-turn or
multi-turn QA exchange.
As shown in \cref{fig:faq_evaluation}, FloCA consistently achieves the highest
accuracy across all compared baselines.
For LLM-based baselines, the current node state is only implicitly represented
in the dialogue history, so intermediate QA exchanges can interfere with the
model's ability to recover the pre-QA flowchart state, leading to incorrect
next-hop transitions.
In contrast, FloCA performs flowchart reasoning through explicit graph
execution, which preserves node state independently of the dialogue context.
This allows FloCA to handle out-of-flowchart knowledge questions while
accurately resuming the interrupted reasoning process.

\begin{figure}[t]
  \centering
  \includegraphics[width=0.85\linewidth]{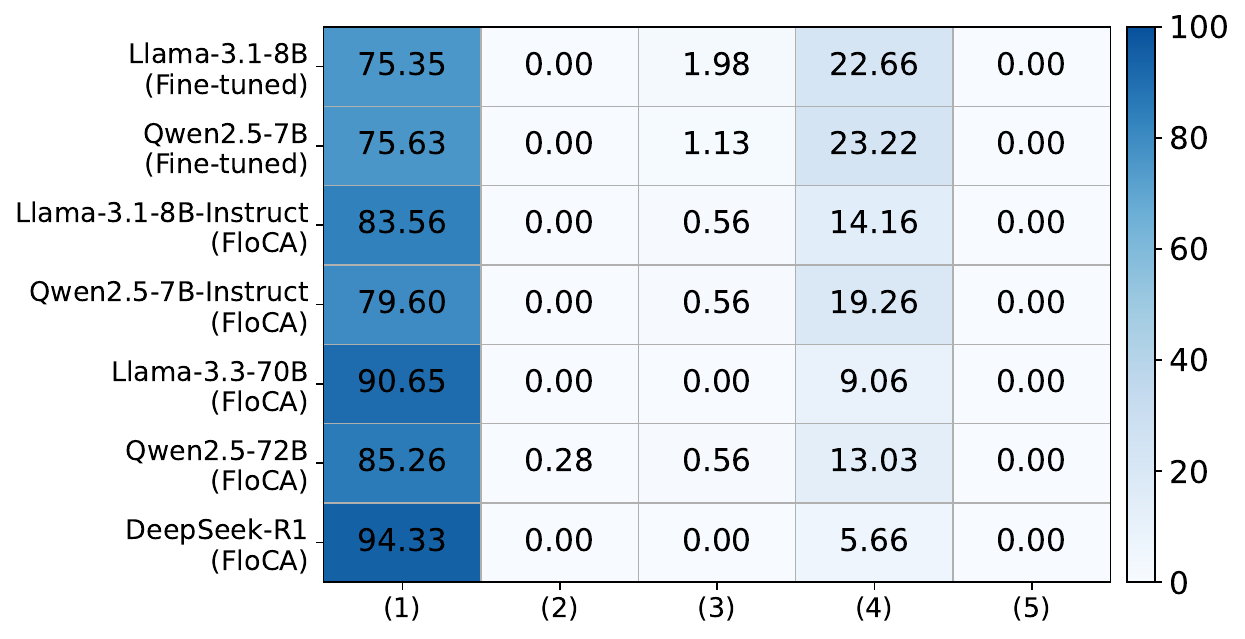}
  \caption{Distribution of path-coverage relations on PFDial in the \emph{in-domain} setting. Each value denotes the percentage of test samples belonging to the corresponding category.}
  \label{fig:path_coverage}
\end{figure}

\subsection{Results on PFDial Dataset}

\begin{table}[t]
\centering
\resizebox{\linewidth}{!}{
\begin{tabular}{lccccc}
\toprule
\multirow{2}{*}{\textbf{Method}}
& \multicolumn{3}{c}{\textbf{Reasoning Metrics}}
& \multicolumn{2}{c}{\textbf{Efficiency Metrics}}\\
\cmidrule(lr){2-4}\cmidrule(lr){5-6}
 & \textbf{INGA}$\uparrow$ & \textbf{TNGA}$\uparrow$ & \textbf{PCA}$\uparrow$ & \textbf{TR}$\downarrow$ & \textbf{NSR}$\downarrow$ \\
\midrule
\rowcolor{gray!15}
\multicolumn{6}{c}{\textbf{\textit{Fine-tuned on PFDial}}}\\
Llama-3.1-8B   & \textbf{100.00} & 92.63 & 75.35 & \textbf{0.00} & \underline{0.77} \\
Qwen2.5-7B     & \textbf{100.00} & 95.18 & 75.63 & \textbf{0.00} & 0.80 \\
\midrule
\rowcolor{gray!15}
\multicolumn{6}{c}{\textbf{\textit{FloCA}}}\\
Llama-3.1-8B-Instruct
& \textbf{100.00} & 92.91 & 83.56 & \textbf{0.00} & 10.66 \\
Qwen2.5-7B-Instruct
& \textbf{100.00} & 94.05 & 79.60 & \textbf{0.00} & 8.33 \\
Llama-3.3-70B
& \textbf{100.00} & \underline{97.73} & \underline{90.65} & \textbf{0.00} & 2.71 \\
Qwen2.5-72B
& \textbf{100.00} & 94.33 & 85.54 & \textbf{0.00} & 2.71\\
\rowcolor[RGB]{234,240,253} DeepSeek-R1
& \textbf{100.00} & \textbf{98.30} & \textbf{94.33} & \textbf{0.00} & \textbf{0.60} \\
\bottomrule
\end{tabular}}
\caption{\emph{In-domain} FOD results on PFDial. All values are multiplied by 100 for better readability. The {\bf bold font} highlights the best results, and the \underline{underline} highlights the second-best results.}
\label{tab:results_on_PFDial_id}
\end{table}

\begin{table}[t]
\centering
\small
\setlength{\tabcolsep}{4pt}
\renewcommand{\arraystretch}{0.9}
\resizebox{\linewidth}{!}{\begin{tabular}{cccc}
\toprule
\textbf{Group} & \textbf{Method} &
\textbf{TNGA (\%)} & \textbf{PCA} (\%) \\
\midrule
1 & w/o self-loop mechanism
  & 84.67 & 83.02 \\
2 & w/o edge matching
  & 91.30 & 90.26 \\
3 & w/o FAQ interruption recovery
  & 90.68 & 89.64\\
\midrule
-- & FloCA
   & \textbf{91.92} & \textbf{90.88} \\
\bottomrule
\end{tabular}
}
\caption{Ablation study of FloCA on FLODIAL using GPT-5.}
\label{tab:ablation}
\end{table}

\Cref{tab:results_on_PFDial_id} compares FloCA with baselines fine-tuned on PlantUML flowcharts in the in-domain setting.
Here we mainly answer the following question: \textbf{Does SFT on serialized flowcharts enable reliable multi-turn flowchart reasoning in LLMs?}
The results suggest that SFT alone is insufficient for reliable multi-turn flowchart reasoning.
Although the fine-tuned baselines achieve TNGA comparable to FloCA, their PCA remains consistently lower than FloCA under similar model sizes. This gap indicates that SFT on serialized flowcharts may improve local semantic matching between user utterances and
flowchart content, but it still fails to reliably maintain topology-consistent node transitions across dialogue turns.
The out-of-domain results in Appendix~\ref{APP:pdfdial_results_ood} further show that SFT generalizes poorly to unseen domains, with performance dropping sharply compared with the in-domain setting.

To further analyze the failure patterns, we categorize each test sample into one
of five path-coverage relations between the ground-truth path and the
predicted path: (1) exact match, (2) ground-truth contained in prediction, (3) prediction contained in
ground-truth, (4) partial overlap without containment, and
(5) disjoint paths.
As shown in \cref{fig:path_coverage}, most samples fall into group (1) or (4), indicating that many failures arise from deviating to an
incorrect branch during intermediate turns. This further supports that serialized-flowchart SFT is better at learning local alignment than globally consistent multi-turn
flowchart reasoning.

\subsection{Ablation Study}
We evaluate the efficacy of three key components in FloCA: the self-loop mechanism, edge matching for interactive flowchart reasoning, and explicit node-state preservation during FAQ interruptions. \Cref{tab:ablation} reports the results.
In ablation 1, the LLM directly predicts the initial node from the serialized flowchart. In ablation 2, the LLM predicts the next node transition based on the current node and its one-hop serialized subgraph. In ablation 3, the LLM recovers the current node state from the dialogue history and serialized flowchart. From the results, all removals degrade TNGA and PCA, with self-loop mechanism removal causing the largest drop.
Instead of directly predicting the initial node from the entire serialized flowchart, it iteratively grounds from the root through topology-constrained execution, reducing ambiguity among semantically similar nodes and preventing topology-inconsistent transitions.

\subsection{Flowchart Reasoning Failure Analysis}
\begin{table}[!t]
  \centering
  \fontsize{8pt}{8pt}\selectfont
  \begin{tblr}{
    width=\linewidth,
    rowsep=1.5pt,
    stretch=0,
    colspec={X[1.25,c]|X[1.8,l]|X[1.8,c]},
    cell{1}{1}={c=2}{halign=c, valign=m},
    cell{2}{1}={c=2}{halign=c, valign=m},
    cell{3}{1}={r=6}{},
    cell{9}{1}={r=6}{},
    cell{15}{1}={r=4}{},
    cell{19}{1}={r=6}{},
    row{1}={font=\bfseries, halign=c, valign=m},
    column{1}={font=\bfseries},
  }
    \toprule
    Method
    &
    & Failure Rate (\%) \\
    \midrule

    SA-FloNet~\cite{raghu2022structural}
    & & 12.21 \\
    \midrule

    Graph Serialization Methods
    & Llama-3.3-70B   & 2.89 \\
    & Claude Opus 4.1 & 2.86 \\
    & GPT-4o          & 0.82 \\
    & GPT-5           & 5.17 \\
    & Qwen3           & 8.48 \\
    & DeepSeek-R1     & 1.65 \\
    \midrule

    RAG-enhanced Graph Serialization Methods
    & Llama-3.3-70B   & 3.51 \\
    & Claude Opus 4.1 & 2.71 \\
    & GPT-4o          & 0.72 \\
    & GPT-5           & 5.17 \\
    & Qwen3           & 4.34 \\
    & DeepSeek-R1     & 1.65 \\
    \midrule

    Visual Language Models
    & Claude Opus 4.1 & 24.22 \\
    & Qwen-VL-Max     & 13.87 \\
    & GPT-4o          & 14.28 \\
    & GPT-5           & 6.00 \\
    \midrule

    FloCA
    & Llama-3.3-70B   & 4.96 \\
    & Claude Opus 4.1 & 2.07 \\
    & GPT-4o          & \textbf{0.62} \\
    & GPT-5           & 1.24 \\
    & Qwen3           & 2.69 \\
    & DeepSeek-R1     & 1.24 \\
    \bottomrule
  \end{tblr}

  \caption{Task failure rates attributable to incorrect initial
node grounding on FLODIAL.}
  \label{tab:initial_grounding_failure}
\end{table}

\begin{table}[t]
  \centering
  \fontsize{8pt}{8pt}\selectfont
  \begin{tblr}{
  width=\linewidth,
  rowsep=1.5pt,
  stretch=0,
  colspec={X[1,c]|X[1.8,l]|X[1.8,c]},
  cell{1}{1}={c=2}{c,m},
  cell{2}{1}={r=6}{},
  row{1}={font=\bfseries, c,m},
  column{1}={font=\bfseries},
}
    \toprule
    Method
    &
    & Failure Rate (\%) \\
    \midrule

    FloCA
    & Llama-3.3-70B   & 16.14 \\
    & Claude Opus 4.1 & 8.69  \\
    & GPT-4o          & 8.28  \\
    & GPT-5           & \textbf{8.07} \\
    & Qwen3           & 8.28  \\
    & DeepSeek-R1     & 9.52  \\

    \bottomrule
  \end{tblr}

  \caption{Task failure rates attributable to LLM
edge matching errors on FLODIAL.}
  \label{tab:edge_matching_failure}
\end{table}
FloCA guarantees structural faithfulness by constraining node transition to the topology. However, semantic reasoning still depends on LLM-based edge matching.
We analyze task failures from two perspectives: the downstream impact of incorrect initial node grounding and the remaining edge-matching errors in FloCA.

\header{Incorrect Initial Node Grounding.}
Incorrect initial grounding may place the dialogue on an
incorrect reasoning trajectory. We regard a task as failed when
incorrect initial grounding results in completely disjoint predicted and ground-truth paths. As shown in
\Cref{tab:initial_grounding_failure}, FloCA with GPT-4o achieves
the lowest failure rate. Instead of inferring the initial node
directly from the global flowchart, FloCA restricts each
semantic decision to the outgoing edge conditions of the current
node, reducing the reasoning burden on the LLM and yielding more
reliable initialization.

\header{Edge-Matching Failures.}
In this analysis, a task is considered failed when the predicted
and ground-truth paths either partially overlap without
containment or are completely disjoint. As shown in \Cref{tab:edge_matching_failure}, the
failure rate decreases from 16.14\% with Llama-3.3-70B to 8.07\%
with GPT-5. This indicates that semantic edge matching remains
dependent on the language understanding capability of the
backbone LLM. The results reveal that even a strong LLM remains sensitive to ambiguous user input and superficial lexicality in binary decision grounding. A detailed error taxonomy with representative examples is provided in \Cref{app:edge_matching_errors}.

\subsection{User Simulator Evaluation}
\begin{table}[t]
\centering
\small
\setlength{\tabcolsep}{6pt}
\renewcommand{\arraystretch}{1.15}
\resizebox{\linewidth}{!}{
\begin{tabular}{lccccc}
\toprule
\textbf{Data Source} & \textbf{SE}$\uparrow$ & \textbf{CE}$\uparrow$ & \textbf{MSTTR}$\uparrow$ & \textbf{MTLD}$\uparrow$ & \textbf{HDD}$\uparrow$ \\
\midrule
\rowcolor{gray!15}
\multicolumn{6}{c}{\textbf{\textit{FLODIAL}}}\\
Annotated Dialogues & 3.61 & 0.13 & 0.91 & 37.09 & 14.48 \\
User Simulator & \textbf{3.79} & \textbf{0.16} & \textbf{0.92} & \textbf{43.57} & \textbf{16.46} \\
\rowcolor{gray!15}
\multicolumn{6}{c}{\textbf{\textit{PFDial}}}\\
Annotated Dialogues & 2.35 & 0.03 & 0.97 & 6.45 & 5.46 \\
User Simulator & \textbf{2.62} & \textbf{0.04} & \textbf{0.98} & \textbf{11.33} & \textbf{7.00} \\
\bottomrule
\end{tabular}
}
\caption{Language diversity comparison of user utterances in the \emph{in-domain} setting. Metrics are computed per dialogue turn and averaged over all test samples.}
\label{tab:lexdiv_turn_level}
\end{table}

We first verify that our user simulator does not introduce factual violations using GPT-5 as an LLM judge. Details are provided in Appendix \ref{App:user_simulator}.
We then evaluate the linguistic diversity of generated user utterances. As shown in \cref{tab:lexdiv_turn_level}, compared to static annotated dialogues, the simulator achieves higher scores across all metrics ~\cite{terragni2023context}, indicating richer surface realizations while preserving the underlying semantics.

\section{Conclusion}

We formally define the FOD task and propose FloCA, a zero-shot conversational agent that achieves faithful flowchart reasoning via a topology-constrained external tool. We further introduce a novel evaluation framework for FOD. Extensive experiments demonstrate that FloCA achieves robust and logically consistent reasoning, outperforming existing LLM/VLM-based methods.

\section*{Limitations}
In this work, we propose a zero-shot conversational agent (FloCA) to address the
FOD task.
We analyze the limitations for future improvement: (1) \emph{Backward transition
  and state correction}.
FloCA currently assumes only forward progression along the flowchart path.
However, real-world scenarios often require backtracking.
For example, in troubleshooting, a user may report an incorrect earlier
judgment, e.g., ``I thought the fuse was blown, but it actually isn't''.
Handling such cases requires the agent to support backward transitions, moving
from the current node to a previous node or a different branch.
(2) \emph{Complex flowchart modeling}.
Our framework models flowcharts as text-attributed graphs, which assumes a
single, fully-connected graph structure.
However, real-world flowcharts often contain more complex elements, such as
swimlanes, disconnected subgraphs, off-page connectors (linking to other
flowcharts or pages).
These irregularities pose challenges for integrating disparate process
information into a unified graph representation.
(3) \emph{Lack of large-scale and real-world datasets}.
While FloCA demonstrates strong performance on existing datasets such as FLODIAL
and PFDial, these datasets are limited in scale and domain coverage.
The scarcity of large-scale, real-world flowchart-oriented dialogue datasets
hinders a more comprehensive evaluation of FOD.

Despite these limitations, FloCA still achieves optimal performance over
existing LLM/VLM-based methods, which can be a strong baseline in future
research.
In future work, we will extend FloCA to more challenging FOD settings, including
backtracking for revisiting earlier decisions and handling irregular flowcharts.

\section*{Acknowledgments}
This work was supported in part by Xi'an Jiaotong University-China Mobile Communications Group Co.,Ltd. Joint Institute and National Natural Science Foundation of China (62272372).

% This document has been adapted
% by Steven Bethard, Ryan Cotterell and Rui Yan
% from the instructions for earlier ACL and NAACL proceedings, including those for
% ACL 2019 by Douwe Kiela and Ivan Vuli\'{c},
% NAACL 2019 by Stephanie Lukin and Alla Roskovskaya,
% ACL 2018 by Shay Cohen, Kevin Gimpel, and Wei Lu,
% NAACL 2018 by Margaret Mitchell and Stephanie Lukin,
% Bib\TeX{} suggestions for (NA)ACL 2017/2018 from Jason Eisner,
% ACL 2017 by Dan Gildea and Min-Yen Kan,
% NAACL 2017 by Margaret Mitchell,
% ACL 2012 by Maggie Li and Michael White,
% ACL 2010 by Jing-Shin Chang and Philipp Koehn,
% ACL 2008 by Johanna D. Moore, Simone Teufel, James Allan, and Sadaoki Furui,
% ACL 2005 by Hwee Tou Ng and Kemal Oflazer,
% ACL 2002 by Eugene Charniak and Dekang Lin,
% and earlier ACL and EACL formats written by several people, including
% John Chen, Henry S. Thompson and Donald Walker.
% Additional elements were taken from the formatting instructions of the \emph{International Joint Conference on Artificial Intelligence} and the \emph{Conference on Computer Vision and Pattern Recognition}.


\bibliography{custom}

\begin{thebibliography}{55}
\providecommand{\natexlab}[1]{#1}

\bibitem[{Abbasian et~al.(2025)Abbasian, Azimi, Rahmani, and
  Jain}]{abbasian2025conversational}
Mahyar Abbasian, Iman Azimi, Amir~M Rahmani, and Ramesh Jain. 2025.
\newblock Conversational health agents: a personalized large language
  model-powered agent framework.
\newblock \emph{JAMIA open}, 8(4):ooaf067.

\bibitem[{Acikgoz et~al.(2025)Acikgoz, Qian, Wang, Dongre, Chen, Ji,
  Hakkani-T{\"u}r, and Tur}]{acikgoz2025desideratum}
Emre~Can Acikgoz, Cheng Qian, Hongru Wang, Vardhan Dongre, Xiusi Chen, Heng Ji,
  Dilek Hakkani-T{\"u}r, and Gokhan Tur. 2025.
\newblock A desideratum for conversational agents: Capabilities, challenges,
  and future directions.
\newblock \emph{arXiv:2504.16939}.

\bibitem[{Baidya et~al.(2025)Baidya, Das, and Gao}]{baidya2025behavior}
Avinash Baidya, Kamalika Das, and Xiang Gao. 2025.
\newblock The behavior gap: Evaluating zero-shot llm agents in complex
  task-oriented dialogs.
\newblock In \emph{Findings of the Association for Computational Linguistics:
  ACL 2025}, pages 23455--23472.

\bibitem[{Bao et~al.(2024)Bao, Peng, Deng, Zhong, Gendron, Pistotti, Tan,
  Young, Chen, Zhu et~al.}]{bao2024abstract}
Qiming Bao, Alex~Yuxuan Peng, Zhenyun Deng, Wanjun Zhong, Gael Gendron, Timothy
  Pistotti, Ne{\c{s}}et Tan, Nathan Young, Yang Chen, Yonghua Zhu, et~al. 2024.
\newblock Abstract meaning representation-based logic-driven data augmentation
  for logical reasoning.
\newblock In \emph{Findings of the Association for Computational Linguistics:
  ACL 2024}, pages 5914--5934.

\bibitem[{Besta et~al.(2024)Besta, Blach, Kubicek, Gerstenberger, Podstawski,
  Gianinazzi, Gajda, Lehmann, Niewiadomski, Nyczyk et~al.}]{besta2024graph}
Maciej Besta, Nils Blach, Ales Kubicek, Robert Gerstenberger, Michal
  Podstawski, Lukas Gianinazzi, Joanna Gajda, Tomasz Lehmann, Hubert
  Niewiadomski, Piotr Nyczyk, et~al. 2024.
\newblock Graph of thoughts: Solving elaborate problems with large language
  models.
\newblock In \emph{Proceedings of the AAAI conference on artificial
  intelligence}.

\bibitem[{Callewaert et~al.(2025)Callewaert, Vandevelde, and
  Vennekens}]{callewaert2025verus}
Benjamin Callewaert, Simon Vandevelde, and Joost Vennekens. 2025.
\newblock Verus-lm: a versatile framework for combining llms with symbolic
  reasoning.
\newblock \emph{arXiv preprint arXiv:2501.14540}.

\bibitem[{Chen et~al.(2025)Chen, Yi, You et~al.}]{chen2025enhancing}
Xi~Chen, Huahui Yi, Mingke You, et~al. 2025.
\newblock Enhancing diagnostic capability with multi-agents conversational
  large language models.
\newblock \emph{NPJ digital medicine}, 8(1):159.

\bibitem[{He et~al.(2025)He, Zeng, Jiang, Zhang, Liu, Shi, and
  Zhou}]{he2025flow2code}
Mengliang He, Jiayi Zeng, Yankai Jiang, Wei Zhang, Zeming Liu, Xiaoming Shi,
  and Aimin Zhou. 2025.
\newblock Flow2code: Evaluating large language models for flowchart-based code
  generation capability.
\newblock In \emph{Findings of the Association for Computational Linguistics:
  ACL 2025}, pages 8124--8146.

\bibitem[{Herrera-Camara and Hammond(2017)}]{herrera2017flow2code}
Jorge-Ivan Herrera-Camara and Tracy Hammond. 2017.
\newblock Flow2code: from hand-drawn flowcharts to code execution.
\newblock In \emph{Proceedings of the Symposium on Sketch-Based Interfaces and
  Modeling}, pages 1--13.

\bibitem[{Huang et~al.(2025)Huang, Lian, Lei et~al.}]{huang2025recommender}
Xu~Huang, Jianxun Lian, Yuxuan Lei, et~al. 2025.
\newblock Recommender ai agent: Integrating large language models for
  interactive recommendations.
\newblock \emph{ACM Transactions on Information Systems}, 43(4):1--33.

\bibitem[{Karimipour et~al.(2025)Karimipour, Pradel, Kellogg, and
  Sridharan}]{karimipour2025llm}
Nima Karimipour, Michael Pradel, Martin Kellogg, and Manu Sridharan. 2025.
\newblock Llm-based repair of static nullability errors.
\newblock \emph{arXiv preprint arXiv:2507.20674}.

\bibitem[{Kim et~al.(2024)Kim, Park, Jeong, Chan, Xu, McDuff, Lee, Ghassemi,
  Breazeal, and Park}]{kim2024mdagents}
Yubin Kim, Chanwoo Park, Hyewon Jeong, Yik~S Chan, Xuhai Xu, Daniel McDuff,
  Hyeonhoon Lee, Marzyeh Ghassemi, Cynthia Breazeal, and Hae~W Park. 2024.
\newblock Mdagents: An adaptive collaboration of llms for medical
  decision-making.
\newblock \emph{Advances in Neural Information Processing Systems},
  37:79410--79452.

\bibitem[{Li et~al.(2023)Li, Meng, Shi, Zhai, and Ruan}]{li2023meddm}
Binbin Li, Tianxin Meng, Xiaoming Shi, Jie Zhai, and Tong Ruan. 2023.
\newblock Meddm: Llm-executable clinical guidance tree for clinical
  decision-making.
\newblock \emph{arXiv preprint arXiv:2312.02441}.

\bibitem[{Li et~al.(2024)Li, Chen, Ross, Huber, Moon, Lin, Dong, Sagar, Yan,
  and Crook}]{li2024large}
Zekun Li, Zhiyu Chen, Mike Ross, Patrick Huber, Seungwhan Moon, Zhaojiang Lin,
  Xin~Luna Dong, Adithya Sagar, Xifeng Yan, and Paul~A Crook. 2024.
\newblock Large language models as zero-shot dialogue state tracker through
  function calling.
\newblock In \emph{Proceedings of the 62nd Annual Meeting of the Association
  for Computational Linguistics (Volume 1: Long Papers)}, pages 8688--8704.

\bibitem[{Liu et~al.(2025)Liu, Xu, Huang, Zeng, Wang, Wang, Yang, and
  Li}]{liu2025logic}
Tongxuan Liu, Wenjiang Xu, Weizhe Huang, Yuting Zeng, Jiaxing Wang, Xingyu
  Wang, Hailong Yang, and Jing Li. 2025.
\newblock Logic-of-thought: Injecting logic into contexts for full reasoning in
  large language models.
\newblock In \emph{Proceedings of the 2025 Conference of the Nations of the
  Americas Chapter of the Association for Computational Linguistics: Human
  Language Technologies}, pages 10168--10185.

\bibitem[{Liu et~al.(2022)Liu, Hu, Zhou, Li, Zhang, and Xiang}]{liu2022code}
Zejie Liu, Xiaoyu Hu, Deyu Zhou, Lin Li, Xu~Zhang, and Yanzheng Xiang. 2022.
\newblock Code generation from flowcharts with texts: A benchmark dataset and
  an approach.
\newblock In \emph{Findings of the Association for Computational Linguistics:
  EMNLP 2022}, pages 6069--6077.

\bibitem[{McLachlan and Webley(2021)}]{mclachlan2021visualisation}
Scott McLachlan and Lisa~C Webley. 2021.
\newblock Visualisation of law and legal process: An opportunity missed.
\newblock \emph{Information Visualization}, 20(2-3):192--204.

\bibitem[{Mentari et~al.(2024)Mentari, Saputra, Syaifudin, Rozi, and
  Nafidiin}]{mentari2024automatic}
Mustika Mentari, Pramana~Yoga Saputra, Yan~Watequlis Syaifudin, Imam~Fahrur
  Rozi, and Naufal Nafidiin. 2024.
\newblock Automatic java code generation system from flowchart.
\newblock In \emph{Proceedings of the 5th Annual Advanced Technology, Applied
  Science, and Engineering Conference (ATASEC) 2023}, page 102. Springer
  Nature.

\bibitem[{Morishita et~al.(2024)Morishita, Morio, Yamaguchi, and
  Sogawa}]{morishita2024enhancing}
Terufumi Morishita, Gaku Morio, Atsuki Yamaguchi, and Yasuhiro Sogawa. 2024.
\newblock Enhancing reasoning capabilities of llms via principled synthetic
  logic corpus.
\newblock \emph{Advances in Neural Information Processing Systems},
  37:73572--73604.

\bibitem[{Nakajima(2023)}]{NakajimaBabyagi2023}
Yohei Nakajima. 2023.
\newblock Babyagi.
\newblock \url{https://github.com/yoheinakajima/babyagi}.

\bibitem[{Onami et~al.(2025)Onami, Miyanishi, Maeda, and
  Kurita}]{onami-etal-2025-legalviz}
Eri Onami, Taiki Miyanishi, Koki Maeda, and Shuhei Kurita. 2025.
\newblock Legalviz: Legal text visualization by text to diagram generation.
\newblock In \emph{Proceedings of the 2025 Conference of the Nations of the
  Americas Chapter of the Association for Computational Linguistics: Human
  Language Technologies (Volume 1: Long Papers)}, pages 6657--6676.

\bibitem[{Pan et~al.(2024)Pan, Zhang, Caragea, Dragut, and
  Jan~Latecki}]{pan2024flowlearn}
Huitong Pan, Qi~Zhang, Cornelia Caragea, Eduard Dragut, and Longin Jan~Latecki.
  2024.
\newblock Flowlearn: Evaluating large vision-language models on flowchart
  understanding.
\newblock In \emph{ECAI 2024}, pages 73--80. IOS Press.

\bibitem[{Pan et~al.(2023)Pan, Albalak, Wang, and Wang}]{pan2023logic}
Liangming Pan, Alon Albalak, Xinyi Wang, and William Wang. 2023.
\newblock Logic-lm: Empowering large language models with symbolic solvers for
  faithful logical reasoning.
\newblock In \emph{Findings of the Association for Computational Linguistics:
  EMNLP 2023}, pages 3806--3824.

\bibitem[{Raghu et~al.(2021)Raghu, Agarwal, Joshi et~al.}]{raghu2021end}
Dinesh Raghu, Shantanu Agarwal, Sachindra Joshi, et~al. 2021.
\newblock End-to-end learning of flowchart grounded task-oriented dialogs.
\newblock In \emph{Proceedings of the 2021 conference on empirical methods in
  natural language processing}, pages 4348--4366.

\bibitem[{Raghu et~al.(2022)Raghu, Joshi, Joshi et~al.}]{raghu2022structural}
Dinesh Raghu, Suraj Joshi, Sachindra Joshi, et~al. 2022.
\newblock Structural constraints and natural language inference for end-to-end
  flowchart grounded dialog response generation.
\newblock In \emph{Proceedings of the 2022 conference on empirical methods in
  natural language processing}, pages 10763--10774.

\bibitem[{ReworkdAI(2025)}]{reworkdAgentGPT2025}
ReworkdAI. 2025.
\newblock Agent{GPT}.
\newblock \url{https://github.com/reworkd/AgentGPT}.

\bibitem[{Richards(2023)}]{RichardsAutogpt2023}
Toran~Bruce Richards. 2023.
\newblock Auto-{GPT}.
\newblock \url{https://github.com/Significant-Gravitas/Auto-GPT}.

\bibitem[{Ryu et~al.(2025)Ryu, Kim, Lee, and Yang}]{ryu2025divide}
Hyun Ryu, Gyeongman Kim, Hyemin~S. Lee, and Eunho Yang. 2025.
\newblock Divide and translate: Compositional first-order logic translation and
  verification for complex logical reasoning.
\newblock In \emph{The Thirteenth International Conference on Learning
  Representations}.

\bibitem[{Schick et~al.(2023)Schick, Dwivedi-Yu, Dess{\`\i}, Raileanu, Lomeli,
  Hambro, Zettlemoyer, Cancedda, and Scialom}]{schick2023toolformer}
Timo Schick, Jane Dwivedi-Yu, Roberto Dess{\`\i}, Roberta Raileanu, Maria
  Lomeli, Eric Hambro, Luke Zettlemoyer, Nicola Cancedda, and Thomas Scialom.
  2023.
\newblock Toolformer: Language models can teach themselves to use tools.
\newblock \emph{Advances in neural information processing systems},
  36:68539--68551.

\bibitem[{Shen et~al.(2024)Shen, Ding, Meng, Xiang et~al.}]{shen2024rethinking}
Lingdong Shen, Kun Ding, Gaofeng Meng, Shiming Xiang, et~al. 2024.
\newblock Rethinking comprehensive benchmark for chart understanding: A
  perspective from scientific literature.
\newblock \emph{arXiv preprint arXiv:2412.12150}.

\bibitem[{Shen et~al.(2023)Shen, Song, Tan, Li, Lu, and
  Zhuang}]{shen2023hugginggpt}
Yongliang Shen, Kaitao Song, Xu~Tan, Dongsheng Li, Weiming Lu, and Yueting
  Zhuang. 2023.
\newblock Hugginggpt: Solving ai tasks with chatgpt and its friends in hugging
  face.
\newblock \emph{Advances in Neural Information Processing Systems},
  36:38154--38180.

\bibitem[{Shinn et~al.(2023)Shinn, Cassano, Gopinath, Narasimhan, and
  Yao}]{shinn2023reflexion}
Noah Shinn, Federico Cassano, Ashwin Gopinath, Karthik Narasimhan, and Shunyu
  Yao. 2023.
\newblock Reflexion: Language agents with verbal reinforcement learning.
\newblock \emph{Advances in neural information processing systems},
  36:8634--8652.

\bibitem[{Shukla et~al.(2023)Shukla, Gatti, Kumar, Yadav, and
  Mishra}]{shukla2023towards}
Shreya Shukla, Prajwal Gatti, Yogesh Kumar, Vikash Yadav, and Anand Mishra.
  2023.
\newblock Towards making flowchart images machine interpretable.
\newblock In \emph{International Conference on Document Analysis and
  Recognition}, pages 505--521. Springer.

\bibitem[{Singh et~al.(2024)Singh, Chaurasia, Varun, Pandya, Gupta, Gupta, and
  Roth}]{singh2024flowvqa}
Shubhankar Singh, Purvi Chaurasia, Yerram Varun, Pranshu Pandya, Vatsal Gupta,
  Vivek Gupta, and Dan Roth. 2024.
\newblock Flowvqa: Mapping multimodal logic in visual question answering with
  flowcharts.
\newblock In \emph{Findings of the Association for Computational Linguistics:
  ACL 2024}, pages 1330--1350.

\bibitem[{Soman et~al.(2025)Soman, Ranjani, Roychowdhury, Sastry, Jain,
  Gangrade, and Khan}]{soman2025graph}
Sumit Soman, HG~Ranjani, Sujoy Roychowdhury, Venkata Dharma Surya~Narayana
  Sastry, Akshat Jain, Pranav Gangrade, and Ayaaz Khan. 2025.
\newblock A graph-based approach for multi-modal question answering from
  flowcharts in telecom documents.
\newblock \emph{arXiv preprint arXiv:2507.22938}.

\bibitem[{Suri et~al.(2025)Suri, Mathur, Lipka, Dernoncourt, Rossi, Gupta, and
  Manocha}]{suri2025follow}
Manan Suri, Puneet Mathur, Nedim Lipka, Franck Dernoncourt, Ryan~A Rossi, Vivek
  Gupta, and Dinesh Manocha. 2025.
\newblock Follow the flow: Fine-grained flowchart attribution with
  neurosymbolic agents.
\newblock In \emph{Proceedings of the 2025 Conference on Empirical Methods in
  Natural Language Processing}, pages 22496--22519.

\bibitem[{Tannert et~al.(2023)Tannert, Feighelstein, Bogojeska, Shtok, Arbelle,
  Staar, Schumann, Kuhn, and Karlinsky}]{tannert2023flowchartqa}
Simon Tannert, Marcelo~G Feighelstein, Jasmina Bogojeska, Joseph Shtok, Assaf
  Arbelle, Peter~WJ Staar, Anika Schumann, Jonas Kuhn, and Leonid Karlinsky.
  2023.
\newblock Flowchartqa: The first large-scale benchmark for reasoning over
  flowcharts.
\newblock In \emph{Proceedings of the 1st Workshop on Linguistic Insights from
  and for Multimodal Language Processing}, pages 34--46.

\bibitem[{Terragni et~al.(2023)Terragni, Filipavicius, Khau, Guedes, Manso, and
  Mathis}]{terragni2023context}
Silvia Terragni, Modestas Filipavicius, Nghia Khau, Bruna Guedes, Andr{\'e}
  Manso, and Roland Mathis. 2023.
\newblock In-context learning user simulators for task-oriented dialog systems.
\newblock \emph{arXiv preprint arXiv:2306.00774}.

\bibitem[{Tu et~al.(2025)Tu, Schaekermann, Palepu, Saab, Freyberg, Tanno, Wang,
  Li, Amin, Cheng et~al.}]{tu2025towards}
Tao Tu, Mike Schaekermann, Anil Palepu, Khaled Saab, Jan Freyberg, Ryutaro
  Tanno, Amy Wang, Brenna Li, Mohamed Amin, Yong Cheng, et~al. 2025.
\newblock Towards conversational diagnostic artificial intelligence.
\newblock \emph{Nature}, 642(8067):442--450.

\bibitem[{Wan et~al.(2024)Wan, Wang, Yang, Yuan, Huang, He, Jiao, and
  Lyu}]{wan2024logicasker}
Yuxuan Wan, Wenxuan Wang, Yiliu Yang, Youliang Yuan, Jen-tse Huang, Pinjia He,
  Wenxiang Jiao, and Michael Lyu. 2024.
\newblock Logicasker: Evaluating and improving the logical reasoning ability of
  large language models.
\newblock In \emph{Proceedings of the 2024 Conference on Empirical Methods in
  Natural Language Processing}, pages 2124--2155.

\bibitem[{Wang et~al.(2024)Wang, Zhao, Shi, Li, and Zeng}]{wang2024novel}
Xingjin Wang, Jiahao Zhao, Jiahui Shi, Linjing Li, and Daniel Zeng. 2024.
\newblock A novel visual-enhanced dual stream long-term decision framework for
  large language model agents.
\newblock In \emph{International Conference on Neural Information Processing},
  pages 332--346. Springer.

\bibitem[{Xu et~al.(2025{\natexlab{a}})Xu, Jian, Zhao, Pang, Zhang, Wang,
  Zhang, Dong, Monteiro, Liu et~al.}]{xu2025graphomni}
Hao Xu, Xiangru Jian, Xinjian Zhao, Wei Pang, Chao Zhang, Suyuchen Wang, Qixin
  Zhang, Zhengyuan Dong, Joao Monteiro, Bang Liu, et~al. 2025{\natexlab{a}}.
\newblock Graphomni: A comprehensive and extendable benchmark framework for
  large language models on graph-theoretic tasks.
\newblock \emph{arXiv preprint arXiv:2504.12764}.

\bibitem[{Xu et~al.(2025{\natexlab{b}})Xu, Mao, Sun, Che, Xu, and
  Huang}]{xu2025agenttod}
Heng-Da Xu, Xian-Ling Mao, Fanshu Sun, Tian-Yi Che, Chun Xu, and Heyan Huang.
  2025{\natexlab{b}}.
\newblock Agenttod: A task-oriented dialogue agent with a flexible and adaptive
  api calling paradigm.
\newblock \emph{ACM Transactions on Information Systems}, 43(5):1--32.

\bibitem[{Xu et~al.(2024)Xu, Mao, Yang, Sun, and Huang}]{xu2024rethinking}
Heng-Da Xu, Xian-Ling Mao, Puhai Yang, Fanshu Sun, and He-Yan Huang. 2024.
\newblock Rethinking task-oriented dialogue systems: From complex modularity to
  zero-shot autonomous agent.
\newblock In \emph{Proceedings of the 62nd Annual Meeting of the Association
  for Computational Linguistics}, pages 2748--2763.

\bibitem[{Xu et~al.(2025{\natexlab{c}})Xu, Dumontier, and Sun}]{xu2025using}
Xingru Xu, Michel Dumontier, and Chang Sun. 2025{\natexlab{c}}.
\newblock Using clinical guidelines, domain ontology, and llms for personalized
  leukemia treatment recommendations.
\newblock In \emph{CEUR Workshop Proceedings}, volume 4001. CEUR-WS.

\bibitem[{Yamanaka et~al.(2025)Yamanaka, Takahashi, and
  Yamashita}]{yamanaka2025flowchart}
Yuuki Yamanaka, Hiroshi Takahashi, and Tomoya Yamashita. 2025.
\newblock Flowchart-based decision making with large language models.
\newblock In \emph{Findings of the Association for Computational Linguistics:
  ACL 2025}, pages 14836--14842.

\bibitem[{Yang et~al.(2024)Yang, Xu, Yao, Rogers, Zhang, Intille, Shara, Gao,
  and Wang}]{yang2024talk2care}
Ziqi Yang, Xuhai Xu, Bingsheng Yao, Ethan Rogers, Shao Zhang, Stephen Intille,
  Nawar Shara, Guodong~Gordon Gao, and Dakuo Wang. 2024.
\newblock Talk2care: An llm-based voice assistant for communication between
  healthcare providers and older adults.
\newblock \emph{Proceedings of the ACM on Interactive, Mobile, Wearable and
  Ubiquitous Technologies}.

\bibitem[{Yao et~al.(2023)Yao, Yu, Zhao, Shafran, Griffiths, Cao, and
  Narasimhan}]{yao2023tree}
Shunyu Yao, Dian Yu, Jeffrey Zhao, Izhak Shafran, Tom Griffiths, Yuan Cao, and
  Karthik Narasimhan. 2023.
\newblock Tree of thoughts: Deliberate problem solving with large language
  models.
\newblock \emph{Advances in neural information processing systems},
  36:11809--11822.

\bibitem[{Yao et~al.(2022)Yao, Zhao, Yu, Du, Shafran, Narasimhan, and
  Cao}]{yao2022react}
Shunyu Yao, Jeffrey Zhao, Dian Yu, Nan Du, Izhak Shafran, Karthik~R Narasimhan,
  and Yuan Cao. 2022.
\newblock React: Synergizing reasoning and acting in language models.
\newblock In \emph{The eleventh international conference on learning
  representations}.

\bibitem[{Ye et~al.(2025)Ye, Dash, Yin, and Wang}]{ye2025beyond}
Junyi Ye, Ankan Dash, Wenpeng Yin, and Guiling Wang. 2025.
\newblock Beyond end-to-end vlms: Leveraging intermediate text representations
  for superior flowchart understanding.
\newblock In \emph{Proceedings of the 2025 Conference of the Nations of the
  Americas Chapter of the Association for Computational Linguistics: Human
  Language Technologies (Volume 1: Long Papers)}, pages 3534--3548.

\bibitem[{Ye et~al.(2024)Ye, Zhang, Wang, Xu, and Zhang}]{ye2024language}
Ruosong Ye, Caiqi Zhang, Runhui Wang, Shuyuan Xu, and Yongfeng Zhang. 2024.
\newblock Language is all a graph needs.
\newblock In \emph{Findings of the association for computational linguistics:
  EACL 2024}, pages 1955--1973.

\bibitem[{Yin et~al.(2025)Yin, Kim, Mathur, Sarker, and Bansal}]{yin2025talk}
Haoteng Yin, Jinha Kim, Prashant Mathur, Krishanu Sarker, and Vidit Bansal.
  2025.
\newblock How to talk to language models: Serialization strategies for
  structured entity matching.
\newblock In \emph{Findings of the Association for Computational Linguistics:
  NAACL 2025}, pages 7836--7850.

\bibitem[{Zeng et~al.(2024)Zeng, Liu, Lu, Wang, Liu, Dong, and
  Tang}]{zeng2024agenttuning}
Aohan Zeng, Mingdao Liu, Rui Lu, Bowen Wang, Xiao Liu, Yuxiao Dong, and Jie
  Tang. 2024.
\newblock Agenttuning: Enabling generalized agent abilities for llms.
\newblock In \emph{Findings of the Association for Computational Linguistics:
  ACL 2024}, pages 3053--3077.

\bibitem[{Zhang et~al.(2024)Zhang, Yao, Liu, Yu, and Wang}]{zhang2024first}
Enming Zhang, Ruobing Yao, Huanyong Liu, Junhui Yu, and Jiale Wang. 2024.
\newblock First multi-dimensional evaluation of flowchart comprehension for
  multimodal large language models.
\newblock \emph{arXiv preprint arXiv:2406.10057}.

\bibitem[{Zhang et~al.(2025)Zhang, Wang, Shen, Yang, Jiang, Wu, Dou, Chen, Xi,
  Zhang et~al.}]{zhang-etal-2025-pfdial}
Ming Zhang, Yuhui Wang, Yujiong Shen, Tingyi Yang, Changhao Jiang, Yilong Wu,
  Shihan Dou, Qinhao Chen, Zhiheng Xi, Zhihao Zhang, et~al. 2025.
\newblock Pfdial: A structured dialogue instruction fine-tuning method based on
  uml flowcharts.
\newblock In \emph{Findings of the Association for Computational Linguistics:
  ACL 2025}, pages 2626--2649.

\end{thebibliography}

\appendix

\section{Appendix}
\subsection{Datasets}\label{App:datastes}
We evaluate the performance of FloCA on FLODIAL and
PFDial. Table \ref{tab:data_statistic} shows a statistic summary of the datasets.

FLODIAL is a flowchart-oriented dialogue dataset containing 12 troubleshooting flowcharts and 2,738 dialogue samples collected via Amazon Mechanical Turk. The flowcharts cover the decision-making process for diagnosing technical issues in cars and laptops.
Each dialogue corresponds to a specific decision path in the flowchart and is annotated with ground-truth node IDs for both agent and user turns.
Since FloCA and all LLM/VLM-based baselines are evaluated in a zero-shot setting and require no training, we report results on the in-domain setting only on FLODIAL. The experimental results show that even when compared against SA-FloNet, which trained on the in-domain split, FloCA consistently achieves better performance across all metrics.

PFDial is a Chinese flowchart-based dialogue instruction dataset. Flowcharts in PFDial are represented in PlantUML. The advantage of PlantUML is that it employs a code-like structured representation utilizing syntax features such as indentation, branching, and loops, making flowchart descriptions both intuitive to read and convenient for program processing. However, PFDial is originally constructed as an instruction corpus, and dialogues are decomposed into atomic units formatted as five-tuples (i.e., flowchart
description, current node, user input, next node, and agent output), rather than providing complete multi-turn dialogues aligned to the corresponding flowchart path. To enable evaluation on our proposed flowchart-oriented dialogue (FOD) task, we first extract all paths from the $\texttt{\textless start\textgreater}$ node to the $\texttt{\textless end\textgreater}$ node based on each PlantUML flowchart by strictly following PlantUML parsing rules. We then use DeepSeek-V3.2 to synthesize a complete dialogue for each extracted path, conditioning on the attributes of all nodes and edges along the path. The synthesized dialogues follow the same format as FLODIAL. Each dialogue turn is annotated with the corresponding ground-truth node ID, providing accessibility for evaluation and further research on FOD task. All synthesized dialogues are released in our open-source repository.

% Requires: \usepackage{booktabs,multirow}
\begin{table*}[t]
\centering
\resizebox{\linewidth}{!}{
\begin{tabular}{llccccccc}
\toprule
\multirow{2}{*}{\textbf{Datasets}} & \multirow{2}{*}{\textbf{Domain}} &
\multirow{2}{*}{\textbf{Train Dialogs}} & \multirow{2}{*}{\textbf{Val Dialogs}} &
\multicolumn{2}{c}{\textbf{Test Dialogs}} &
\multirow{2}{*}{\textbf{Train flowcharts}} & \multirow{2}{*}{\textbf{Val flowcharts}} & \multirow{2}{*}{\textbf{Test flowcharts}} \\
\cmidrule(lr){5-6}
& & & & \textbf{Root-init} & \textbf{Middle-init} & & & \\
\midrule

\multirow{4}{*}{\textsc{FLODIAL}}
& \multirow{2}{*}{In-domain}    & \multirow{2}{*}{1,798} & \multirow{2}{*}{456}
& Root-init   & Middle-init
& \multirow{2}{*}{12} & \multirow{2}{*}{12} & \multirow{2}{*}{12} \\
&                            &                      &
& 350         & 134
&                    &                    &                    \\
& \multirow{2}{*}{Out-of-domain} & \multirow{2}{*}{1,786} & \multirow{2}{*}{454}
& root node   & middle node
& \multirow{2}{*}{8}  & \multirow{2}{*}{2}  & \multirow{2}{*}{2}  \\
&                            &                      &
& 408         & 90
&                    &                    &                    \\

\midrule

\multirow{4}{*}{\textsc{PFDial}}
& \multirow{2}{*}{In-domain}    & \multirow{4}{*}{12,705} & \multirow{4}{*}{--}
& Root-init   & Middle-init
& \multirow{4}{*}{440} & \multirow{4}{*}{--} & \multirow{2}{*}{80} \\
&                            &                       &
& 513         & 0
&                     &                     &                    \\
& \multirow{2}{*}{Out-of-domain} &                       &
& Root-init   & Middle-init
&                     &                     & \multirow{2}{*}{80} \\
&                            &                       &
& 416         & 0
&                     &                     &                    \\

\bottomrule
\end{tabular}}
\caption{A statistic summary of the datasets.}
\label{tab:data_statistic}
\end{table*}

Table \ref{tab:results_on_PFDial_ood} shows the result of \emph{out-of-domain} FOD results on PFDial. Compared to \emph{in-domain}, the limitations of SFT-based learning become pronounced under the \emph{out-of-domain} setting. For Llama-3.1-8B, all reasoning metrics collapse, accompanied by a high failure rate, which is $48.31\%$.  We define ``Failure'' as cases where the model outputs an empty response or
produces a next-state text that cannot be matched to any node in the flowchart.
The high failure rate shows that when both the flowchart content and user
expressions shift in \emph{out-of-domain}, the SFT model often fails to align an unseen user utterance with the specific node texts of an unseen PlantUML flowchart and thus
generates a non-existent next-state that cannot be grounded to any node in the flowchart.

In contrast, FloCA remains stable across domains with zero Failure for all
backbones. This robustness stems from it decoupling complex flowchart reasoning to an
external tool that executes the graph and enforces topology-valid transitions,
making the process less dependent on fitting a particular training distribution.
These results suggest that robust performance on FOD relies more on
topology-constrained execution than on fitting a training distribution via SFT
on serialized flowcharts.
Although FloCA performs lower on the out-of-domain than in-domain, this
drop is because there are 1,961 decision branches in \emph{out-of-domain} flowcharts, while only 870 in \emph{in-domain}.
The resulting decision space is more than twice as large, increasing ambiguity
at each transition.

\begin{figure}[t]
  \centering
  \includegraphics[width=0.8\linewidth]{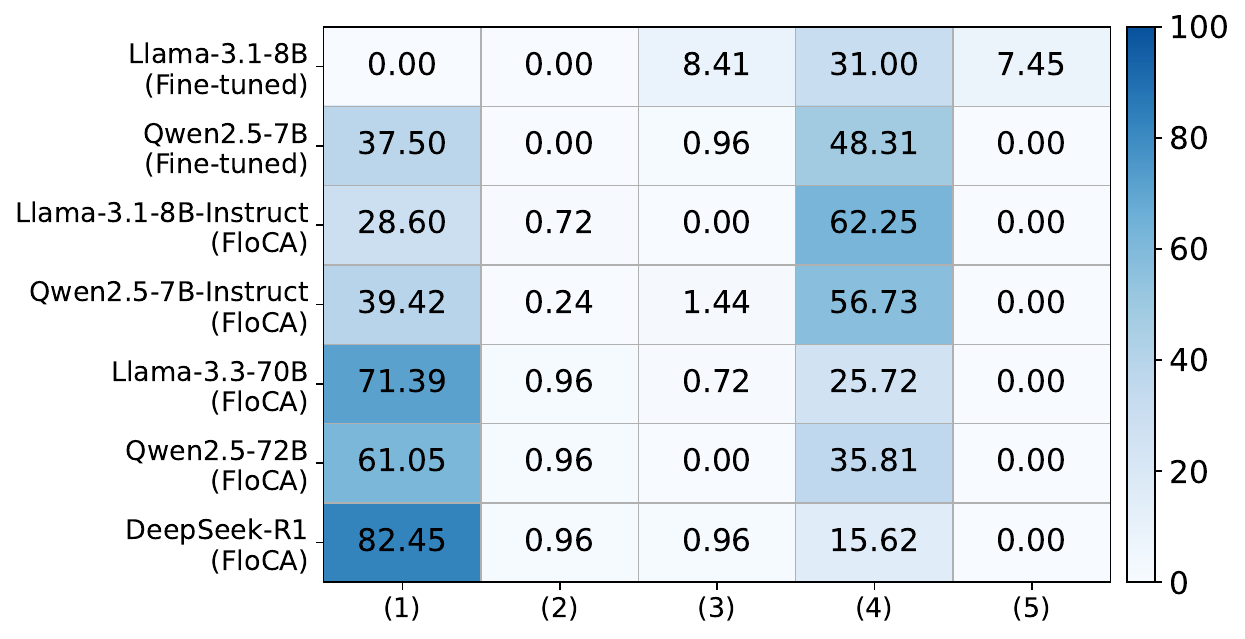}
  \caption{Distribution of path-coverage relations on the PFDial in the \emph{out-of-domain} setting.}
  \label{fig:path_coverage_ood}
  \vspace{-0.5ex}
\end{figure}

To analyze error patterns, we categorize each test sample into one of five path-coverage relations between the ground-truth path and the predicted path:  (1) exact match, (2) ground-truth contained in prediction, (3) prediction contained in
ground-truth, (4) partial overlap without containment, and (5) disjoint paths.
\Cref{fig:path_coverage_ood} shows the path-coverage relations on \emph{out-of-domain} settings. The error pattern is identical with \emph{in-domain} settings, as there are proportional samples that fall into group (4). This indicates that many dialogues deviate from the ground-truth path at intermediate nodes. Therefore, improving overall flowchart reasoning success relies on increasing the correctness of every decision branch selection.

\begin{table}[t]
\centering
\resizebox{\linewidth}{!}{
\begin{tabular}{lcccccc}
\toprule
\multirow{2}{*}{\textbf{Method}}
& \multicolumn{3}{c}{\textbf{Reasoning Metrics}}
& \multicolumn{2}{c}{\textbf{Efficiency Metrics}}
& \multirow{2}{*}{\textbf{Failure}$\downarrow$}\\
\cmidrule(lr){2-4}\cmidrule(lr){5-6}
 & \textbf{INGA}$\uparrow$ & \textbf{TNGA}$\uparrow$ & \textbf{PCA}$\uparrow$ & \textbf{TR}$\downarrow$ & \textbf{NSR}$\downarrow$ \\
\midrule
\rowcolor{gray!15}
\multicolumn{7}{c}{\textbf{\textit{Fine-tuned on PFDial}}}\\
Llama-3.1-8B
& 8.17 & 2.88 & 0.00 & 53.12 & 26.52 & 48.31 \\
Qwen2.5-7B
& \underline{99.27} & 66.34 & 37.50 & 13.22 & 3.66 & \underline{0.72}\\
\midrule
\rowcolor{gray!15}
\multicolumn{7}{c}{\textbf{\textit{FloCA}}}\\
Llama-3.1-8B-Instruct
& \textbf{100.00} & 70.19 & 29.32 & \textbf{0.00} & 4.86 & \textbf{0.00}\\
Qwen2.5-7B-Instruct
& \textbf{100.00} & 65.14 & 39.66 & \textbf{0.00} & 2.53 & \textbf{0.00}\\
Llama-3.3-70B
& \textbf{100.00} & \underline{92.06} & \underline{72.35} & \textbf{0.00} & 2.75 & \textbf{0.00}\\
Qwen2.5-72B
& \textbf{100.00} & 82.69 & 62.01 & \textbf{0.00} & \underline{1.21}& \textbf{0.00} \\
\rowcolor[RGB]{234,240,253} DeepSeek-R1
& \textbf{100.00} & \textbf{97.83} & \textbf{83.41} & \textbf{0.00} & \textbf{1.02} & \textbf{0.00} \\
\bottomrule
\end{tabular}}
\caption{\emph{Out-of-domain} FOD results on PFDial. The {\bf bold font} highlights the best results, and the \underline{underline} highlights the second-best results.}
\label{tab:results_on_PFDial_ood}
\end{table}

\subsection{LLM edge matching error taxonomy}\label{app:edge_matching_errors}
\begin{table*}[t]
  \centering
  \footnotesize
  \begin{tblr}{
    rowsep=2pt,
    stretch=0,
    colspec={
      X[1.15,l]
      X[3.0,l]
      X[.65,c]
      X[3.6,l]
    },
    row{1}={font=\bfseries, c, m},
    cells={m},
  }
    \toprule
    Error Type
    & Description
    & Proportion (\%)
    & Examples \\
    \midrule

    Ambiguous or indirect user response
    &
    The user utterance is ambiguous, incomplete, or does not
    directly answer the agent's question.
    &
    40.0
    &
    \textbf{Agent:} ``Can you check if a fusible link has
    blown?''; \textbf{User:} ``Yes, I can check that. I've done
    it before on this vehicle.''; \textbf{Predicted edge:}
    ``yes''; \textbf{Ground truth:} ``no''
    \\

    \midrule

    Partial semantic understanding
    &
    The user utterance contains sufficient information for
    correct edge matching, but the LLM fails to correctly
    interpret semantic cues such as negation, polarity, or
    logical consistency across multiple sentences.
    &
    48.0
    &
    \textbf{Agent:} ``Can you check if a fusible link has
    blown?''; \textbf{User:} ``There does not appear to be a
    blown fusible link. The fusible link appears okay.'';
    \textbf{Predicted edge:} ``yes''; \textbf{Ground truth:}
    ``no''
    \\

    \midrule

    Negative-question polarity error
    &
    The LLM incorrectly maps the response polarity because it
    over-relies on surface negation words (e.g., ``not'',
    ``nothing'', and ``do not'').
    &
    12.0
    &
    \textbf{Agent:} ``When your Windows boots, is nothing being
    displayed on the LCD?''; \textbf{User:} ``I do not see
    anything on it.''; \textbf{Predicted edge:} ``no'';
    \textbf{Ground truth:} ``yes''
    \\

    \bottomrule
  \end{tblr}

  \caption{Taxonomy of LLM edge matching errors made
    by FloCA with GPT-5 on FLODIAL. The proportions are computed
    over all identified edge-matching failures.}
  \label{tab:edge_matching_taxonomy}
\end{table*}
To analyze the specific error patterns in edge matching, we manually categorize LLM edge matching failures into three representative types, as shown in \Cref{tab:edge_matching_taxonomy}. The results reveal that even a strong LLM remains sensitive to ambiguous user input and superficial lexicality in binary decision grounding.

\subsection{Additional Results on User Simulator}
\begin{table}[t]
\centering
\small
\setlength{\tabcolsep}{6pt}
\renewcommand{\arraystretch}{1.15}
\resizebox{\linewidth}{!}{
\begin{tabular}{lccccc}
\toprule
\textbf{Data Source} & \textbf{SE}$\uparrow$ & \textbf{CE}$\uparrow$ & \textbf{MSTTR}$\uparrow$ & \textbf{MTLD}$\uparrow$ & \textbf{HDD}$\uparrow$ \\
\midrule
\rowcolor{gray!15}
\multicolumn{6}{c}{\textbf{\textit{PFDial}}}\\
Annotated Dialogues & 2.35 & 0.03 & 0.97 & 6.45 & 5.46 \\
User Simulator & \textbf{2.62} & \textbf{0.04} & \textbf{0.98} & \textbf{11.33} & \textbf{7.00} \\
\bottomrule
\end{tabular}
}
\caption{Language diversity comparison of user utterances in the \emph{out-of-domain} setting. Metrics are computed per dialogue turn and averaged over all test samples.}
\label{tab:lexdiv_turn_level_PFDial_ood}
\end{table}

Table \ref{tab:lexdiv_turn_level_PFDial_ood} shows the language diversity evaluation of the user simulator on PFDial in \emph{an out-of-domain} setting.
Metrics include Shannon Entropy (SE), conditional bigram entropy (CE), Mean
Segmental Type–Token Ratio (MSTTR), Measure of Textual Lexical Diversity (MTLD),
and Hyper-geometric Distribution (HDD).
The results show that our proposed user simulator consistently achieves higher
diversity metrics than the static annotated dialogues across all experimented datasets.

\subsection{The Prompt for User Simulator}\label{App:user_simulator}
We first use GPT-5 to extract user-known facts from each annotated dialogue sample. The extraction prompt constrains the LLM to retain only information that would reasonably be available to the user, including the user's goal, observable situation, and previously completed actions or checks.
To simulate scenarios where the dialogue starts from an intermediate flowchart node rather than the root, we augment such samples with extra user-known information. Specifically, for each sample whose annotated initial node is not the root, we extract the textual attributes along the flowchart path from the root node to the annotated initial node and provide them in the prompt as ``Pre-established actions''. This information represents actions or checks that the user would already have completed or known.
The prompt of extraction is presented in \Cref{fig:prompt_template_known_fact_extract}.

We then use DeepSeek-R1 to construct the user simulator in our experiments. The prompt template is presented in \Cref{fig:prompt_template_user_simulator}. For each test dialogue, we initialize the interaction with the ground-truth first user utterance. After the agent generates the first response, the user simulator takes over and dynamically generates user replies for all subsequent turns, conditioned on the dialogue history, while ensuring that the generated utterances remain consistent with the facts specified in the annotated dialogue.

\subsection{Additional Results on PFDial Dataset}\label{APP:pdfdial_results_ood}

\subsection{The Prompt for LLM Judge}
As user responses directly affect the agent’s flow reasoning accuracy, it is crucial to ensure that the user simulator does not introduce factually inconsistent replies that could erroneously mislead the agent. To this end, we employ GPT-5 as an LLM judge to verify whether any simulated user utterance in the generated dialogues violates the facts stated in the annotated dialogue. \Cref{fig:factual_judge_prompt} presents the prompt template used by the judge. Specifically, we treat the annotated user utterances as the set of known facts and ask the judge to determine whether each user utterance in the predicted dialogue contradicts any of these facts. For example, if the annotated dialogue states ``my battery is short-circuited,'' but the user simulator claims ``my battery is not short-circuited,'' this will be identified as a factual VIOLATION. From the experimental results, the judge detects no fact violations across all experiments, indicating that the user simulator remains faithful to the ground-truth facts throughout the interaction and ensuring a fair comparison among models.

\subsection{The Prompt for Interactive Flowchart Reasoning}
We use LLM to match each user input to a valid outgoing edge for the next node transition. \Cref{fig:prompt_template_flowchart_reasoning} shows the prompt template for node transition in interactive flowchart reasoning on FLODIAL. Since FLODIAL is built on decision-making flowcharts where each node poses a binary decision question and has two outgoing branches corresponding to “yes” and “no”, the LLM is instructed to classify the user’s utterance as supporting one of these two conditions given the decision question of the current node. FloCA then performs topology-constrained graph execution and moves to the next-hop node according to the predicted edge condition.

\subsection{Case Study}
Below we show the annotated dialogue alongside the dialogues generated by FloCA and three baselines: graph serialization, RAG-enhanced graph serialization, and visual language models. All methods use GPT-5 as the backbone. In this comparison, we regard the annotated dialogue as ground truth.

The results indicate that FloCA can perform faithful flow reasoning by executing topologically valid node transitions, thus ultimately reaching the correct terminal node. We observe that the ground-truth dialogue is annotated to start at node 8, whereas FloCA initializes the dialogue state at the root node. This is because FloCA adopts a self-loop mechanism to automatically perform flowchart reasoning starting from the root, but the initial user utterance provides insufficient evidence to determine a reliable next hop. After several interaction turns with the user simulator, FloCA correctly grounds the dialogue to node 8 and subsequently follows the correct decision path to the terminal node.
In contrast, the \emph{graph serialization method} fails at initial node grounding and then deviates to a wrong path, resulting in the task failure. Notably, the \emph{RAG-enhanced graph serialization} stays at node 12 and spends 2 dialogue turns to advance the flowchart reasoning. This observation suggests that when the retrieved context is misaligned with the model’s current reasoning, it may introduce additional uncertainty. LLM may be distracted from making the correct node transition, thereby increasing NSR metric. Besides, the VLM baseline exhibits severe repetitive clarification. It loops at node 12 for eight consecutive dialogue turns and then also transits to a wrong terminal node. This observation is consistent with the NSR results, indicating that vision-based flowchart reasoning is more prone to repeated clarification turns, which increases redundancy and often prevents the dialogue from progressing along the decision path.

The observations in this case study are consistent with the evaluation results. It highlights the limitations of probabilistic generation when performing flowchart reasoning and provides evidence that the explicit topological constraints in FloCA are critical for achieving faithful flowchart reasoning.

\clearpage
\onecolumn

\FloatBarrier
\begin{tcolorbox}[
  label=case:groundtruth,
  title={Case Study: Ground-truth Dialogue},
  colback=white,
  colframe=black,
  boxrule=0.8pt,
  arc=1.5mm,
  breakable,
  enhanced,
  left=1.2mm,right=1.2mm,top=1.0mm,bottom=1.0mm
]
\begin{lstlisting}[style=casejson]
{"utterance": [
    {
      "speaker": "user",
      "utterance": "I have a Dell laptop and its fan never runs. It has started to overheat really badly and I am quite worried about it. Can you help me figure this out?",
      "turn_id": 1
    },
    {
      "speaker": "agent",
      "utterance": "Did you put your laptop on a flat and hard surface?",
      "turn_id": 1,
@@      "pre_doc_id": "chart-8"@@
    },
    {
      "speaker": "user",
      "utterance": "I have the laptop on a table.",
      "turn_id": 2
    },
    {
      "speaker": "agent",
      "utterance": "Does the overheating cause your laptop to shut down?",
      "turn_id": 2,
@@      "pre_doc_id": "chart-11"@@
    },
    {
      "speaker": "user",
      "utterance": "It shuts down when it overheats.",
      "turn_id": 3
    },
    {
      "speaker": "agent",
      "utterance": "Do you have BIOS set to put the laptop into hibernation upon low battery?",
      "turn_id": 3,
@@      "pre_doc_id": "chart-14"@@
    },
    {
      "speaker": "user",
      "utterance": "There is no such setting. My laptop continues to run all the time.",
      "turn_id": 4
    },
    {
      "speaker": "agent",
      "utterance": "I recognize the problem with your laptop now. You need to clean all the air vents and replace the active heat sink.",
      "turn_id": 4,
@@      "pre_doc_id": "chart-26"@@
    }
  ],
  "flowchart": "laptop_overheating"}
\end{lstlisting}
\end{tcolorbox}
\FloatBarrier

\begin{tcolorbox}[
  title={Case Study: Predicted Dialogue Generated by FloCA with GPT-5},
  colback=white,
  colframe=black,
  boxrule=0.8pt,
  arc=1.5mm,
  breakable,
  enhanced,
  left=1.2mm,right=1.2mm,top=1.0mm,bottom=1.0mm
]
\begin{lstlisting}[style=casejson]
{
  "utterance": [
    {
      "speaker": "user",
      "utterance": "I have a Dell laptop and it's fan never runs. It has started to overheat really bad and I am quite worried about it. Can you help me figure this out?",
      "turn_id": 1
    },
    {
      "speaker": "agent",
      "utterance": "Do you smell burning or see any smoke coming from the laptop?",
      "turn_id": 1,
@@      "pre_doc_id": "chart-1"@@
    },
    {
      "speaker": "user",
      "utterance": "No, I don't smell any burning or see smoke coming from the laptop. It's just getting extremely hot to the touch.",
      "turn_id": 2
    },
    {
      "speaker": "agent",
      "utterance": "Are you using your laptop in a hot environment?",
      "turn_id": 2,
@@      "pre_doc_id": "chart-2"@@
    },
    {
      "speaker": "user",
      "utterance": "No, I'm using it indoors at normal room temperature.",
      "turn_id": 3
    },
    {
      "speaker": "agent",
      "utterance": "Please tell me if your laptop's battery is overheating?",
      "turn_id": 3,
@@      "pre_doc_id": "chart-3"@@
    },
    {
      "speaker": "user",
      "utterance": "No, the battery itself doesn't seem to be overheating. It's the main body of the laptop, near the vents and the bottom, that gets extremely hot.",
      "turn_id": 4
    },
    {
      "speaker": "agent",
      "utterance": "Is your laptop's fan running when it overheats?",
      "turn_id": 4,
@@      "pre_doc_id": "chart-6"@@
    },
    {
      "speaker": "user",
      "utterance": "Yes, the fan is running when it overheats. I can definitely hear it.",
      "turn_id": 5
    },
    {
      "speaker": "agent",
      "utterance": "Are you using your laptop on a hard, flat surface?",
      "turn_id": 5,
@@      "pre_doc_id": "chart-8"@@
    },
    {
      "speaker": "user",
      "utterance": "I have the laptop on a table.",
      "turn_id": 6
    },
    {
      "speaker": "agent",
      "utterance": "Does your laptop shut down automatically when it overheats?",
      "turn_id": 6,
@@      "pre_doc_id": "chart-11"@@
    },
    {
      "speaker": "user",
      "utterance": "Yes, it does shut down automatically when it overheats, but only when I'm running something demanding like a game or video editing software.",
      "turn_id": 7
    },
    {
      "speaker": "agent",
      "utterance": "Is your BIOS configured to hibernate the laptop in case of low battery?",
      "turn_id": 7,
@@      "pre_doc_id": "chart-14"@@
    },
    {
      "speaker": "user",
      "utterance": "There is no such setting. My laptop continues to run all the time.",
      "turn_id": 8
    },
    {
      "speaker": "agent",
      "utterance": "I understand the issue now. You will need to clean all the air vents on your laptop and replace the active heat sink.",
      "turn_id": 8,
@@      "pre_doc_id": "chart-26"@@
    }
  ],
  "flowchart": "laptop_overheating"
}
\end{lstlisting}
\end{tcolorbox}

\FloatBarrier
\begin{tcolorbox}[
  title={Case Study: Predicted Dialogue Generated by Graph Serialization Method with GPT-5},
  colback=white,
  colframe=black,
  boxrule=0.8pt,
  arc=1.5mm,
  breakable,
  enhanced,
  left=1.2mm,right=1.2mm,top=1.0mm,bottom=1.0mm
]
\begin{lstlisting}[style=casejson]
{
  "utterances": [
    {
      "speaker": "user",
      "utterance": "I have a Dell laptop and it's fan never runs. It has started to overheat really bad and I am quite worried about it. Can you help me figure this out?",
      "turn_id": 1
    },
    {
      "speaker": "agent",
      "utterance": "Have you recently upgraded your laptop's BIOS?",
      "turn_id": 1,
@@    "pre_doc_id": "chart-7"@@
    },
    {
      "speaker": "user",
      "utterance": "No, I haven't.",
      "turn_id": 2
    },
    {
      "speaker": "agent",
      "utterance": "Have you changed the laptop's power management settings recently?",
      "turn_id": 2,
@@    "pre_doc_id": "chart-12"@@
    },
    {
      "speaker": "user",
      "utterance": "No, I haven't changed any power settings.",
      "turn_id": 3
    },
    {
      "speaker": "agent",
      "utterance": "You're facing overheating due to a hardware failure. Your laptop's fan or temperature sensor is not working properly, and you should replace it.",
      "turn_id": 3,
@@    "pre_doc_id": "chart-19"@@
    }
  ],
  "flowchart": "laptop_overheating"
}
\end{lstlisting}
\end{tcolorbox}
\FloatBarrier

\lstdefinestyle{casejson}{
  basicstyle=\ttfamily\footnotesize,
  columns=fullflexible,
  breaklines=true,
  breakatwhitespace=true,
  keepspaces=true,
  showstringspaces=false,
  tabsize=2,
  moredelim=**[is][\color{red}]{@@}{@@},
}

% ---------- In document ----------
\FloatBarrier
\begin{tcolorbox}[
  title={Case Study: Predicted Dialogue Generated by RAG-enhanced Graph Serialization Method with GPT-5},
  colback=white,
  colframe=black,
  boxrule=0.8pt,
  arc=1.5mm,
  breakable,
  enhanced,
  left=1.2mm,right=1.2mm,top=1.0mm,bottom=1.0mm
]
\begin{lstlisting}[style=casejson]
{
  "utterances": [
    {
      "speaker": "user",
      "utterance": "I have a Dell laptop and it's fan never runs. It has started to overheat really bad and I am quite worried about it. Can you help me figure this out?",
      "turn_id": 1
    },
    {
      "speaker": "agent",
      "utterance": "Have you recently upgraded your laptop's BIOS?",
      "turn_id": 1,
@@    "pre_doc_id": "chart-7"@@
    },
    {
      "speaker": "user",
      "utterance": "No, I haven't done that.",
      "turn_id": 2
    },
    {
      "speaker": "agent",
      "utterance": "Have you changed the laptop's power management settings recently?",
      "turn_id": 2,
@@    "pre_doc_id": "chart-12"@@
    },
    {
      "speaker": "user",
      "utterance": "No, I haven't done that.",
      "turn_id": 2
    },
    {
      "speaker": "agent",
      "utterance": "Have you changed the laptop's power management settings recently?",
      "turn_id": 2,
@@    "pre_doc_id": "chart-12"@@
    },
    {
      "speaker": "user",
      "utterance": "No, I haven't changed any power settings recently.",
      "turn_id": 3
    },
    {
      "speaker": "agent",
      "utterance": "You are experiencing overheating due to a hardware failure. Your laptop's fan or temperature sensor is not working properly and you should replace it.",
      "turn_id": 3,
@@    "pre_doc_id": "chart-19"@@
    },
    }
  ],
  "flowchart": "laptop_overheating"
}
\end{lstlisting}
\end{tcolorbox}
\FloatBarrier

\FloatBarrier
\begin{tcolorbox}[
  title={Case Study: Predicted Dialogue Generated by Visual Language Models with GPT-5},
  colback=white,
  colframe=black,
  boxrule=0.8pt,
  arc=1.5mm,
  breakable,
  enhanced,
  left=1.2mm,right=1.2mm,top=1.0mm,bottom=1.0mm
]
\begin{lstlisting}[style=casejson]
{
  "utterances": [
    {
      "speaker": "user",
      "utterance": "I have a Dell laptop and it's fan never runs. It has started to overheat really bad and I am quite worried about it. Can you help me figure this out?",
      "turn_id": 1
    },
    {
      "speaker": "agent",
      "utterance": "Have you recently upgraded your laptop's BIOS?",
      "turn_id": 1,
@@    "pre_doc_id": "chart-7"@@
    },
    {
      "speaker": "user",
      "utterance": "No, I haven't.",
      "turn_id": 2
    },
    {
      "speaker": "agent",
      "utterance": "Have you changed the laptop's power management settings recently?",
      "turn_id": 2,
@@    "pre_doc_id": "chart-12"@@
    },
    {
      "speaker": "user",
      "utterance": "No, I haven't changed any power settings recently.",
      "turn_id": 3
    },
    {
      "speaker": "agent",
      "utterance": "Did you change the laptop power management settings recently?",
      "turn_id": 3,
@@    "pre_doc_id": "chart-12"@@
    },
    {
      "speaker": "user",
      "utterance": "No, I haven't changed any power settings recently.",
      "turn_id": 4
    },
    {
      "speaker": "agent",
      "utterance": "Have you changed your laptop's power management settings recently?",
      "turn_id": 4,
@@    "pre_doc_id": "chart-12"@@
    },
    {
      "speaker": "user",
      "utterance": "No, I haven't changed any power settings recently.",
      "turn_id": 5
    },
    {
      "speaker": "agent",
      "utterance": "Have you changed the laptop power management settings recently?",
      "turn_id": 5,
@@    "pre_doc_id": "chart-12"@@
    },
    {
      "speaker": "user",
      "utterance": "No, I haven't made any changes to the power management settings.",
      "turn_id": 6
    },
    {
      "speaker": "agent",
      "utterance": "Have you changed the laptop's power management settings recently?",
      "turn_id": 6,
@@    "pre_doc_id": "chart-12"@@
    },
    {
      "speaker": "user",
      "utterance": "No, I haven't made any changes to the power management settings.",
      "turn_id": 7
    },
    {
      "speaker": "agent",
      "utterance": "Have you changed the laptop power management settings recently?",
      "turn_id": 7,
@@    "pre_doc_id": "chart-12"@@
    },
    {
      "speaker": "user",
      "utterance": "No, I haven't changed the power management settings.",
      "turn_id": 8
    },
    {
      "speaker": "agent",
      "utterance": "Did you change the laptop's power management settings recently?",
      "turn_id": 8,
@@    "pre_doc_id": "chart-12"@@
    },
    {
      "speaker": "user",
      "utterance": "No, I haven't changed the power management settings.",
      "turn_id": 9
    },
    {
      "speaker": "agent",
      "utterance": "Have you changed the laptop's power management settings recently?",
      "turn_id": 9,
@@    "pre_doc_id": "chart-12"@@
    },
    {
      "speaker": "user",
      "utterance": "No, I haven't changed any power management settings.",
      "turn_id": 10
    },
    {
      "speaker": "agent",
      "utterance": "Revert your power management settings to their previous state. This should prevent your laptop from overheating.",
      "turn_id": 10,
@@    "pre_doc_id": "chart-15"@@
    }
  ],
  "flowchart": "laptop_overheating"
}
\end{lstlisting}
\end{tcolorbox}
\FloatBarrier

\begin{figure*}[t]
\centering
\begin{promptbox}
\ttfamily\small
\setlength{\parindent}{0pt}
\setlength{\parskip}{0pt}

% label width
\newcommand{\labw}{0.18\textwidth}
\newcommand{\txtw}{\dimexpr\textwidth-\labw\relax}

\noindent\makebox[\labw][l]{\promptlabel{System Prompt:}}%
\begin{minipage}[t]{\txtw}
You are an information extraction assistant.
\end{minipage}

\vspace{0.6ex}

\noindent\makebox[\labw][l]{\promptlabel{User Prompt:}}%
\begin{minipage}[t]{\txtw}
Extraction goal:
Extract only information that the user can reasonably know or observe based on the given annotated dialogue, so that a simulator can consistently play the
role of that user in a multi-turn dialogue.
Given an annotated dialogue sample, extract the following types of information:
\begin{enumerate}[leftmargin=*, topsep=0pt, itemsep=0pt, parsep=0pt, partopsep=0pt]
\item User goal: What the user is trying to solve, achieve, or ask about.
\item Observable situation: troubles, conditions, status, or other observable facts described by the user.
\item Pre-established actions or checks: Actions, inspections, or operations that have already been completed before the conversation begins.
\end{enumerate}

\medskip
Important rules:
\begin{enumerate}[leftmargin=*, topsep=0pt, itemsep=0pt, parsep=0pt, partopsep=0pt]
\item Extract only user-side observable or established facts.
\item Do not infer hidden causes, correct diagnoses, or future outcomes.
\item Do not add facts that are not grounded in the annotated dialogue.
\item Keep the extracted facts concise, neutral, and faithful to the dialogue.
\item Organize the output into the required schema.
\end{enumerate}

\medskip
Output format:\\
\texttt{
\{user\_goal: [...],\\
  observable\_situation: [...],\\
  completed\_actions\_or\_checks: [...]\}
}
\medskip

The given annotated dialogue is below: \\
Annotated dialogue: \texttt{\{dialogue\}}\\
Pre-established actions: \texttt{\{actions\}}
\end{minipage}

\end{promptbox}
\caption{The prompt template for user-known fact extraction.}
\label{fig:prompt_template_known_fact_extract}
\end{figure*}

\begin{figure*}[t]
\centering
\begin{promptbox}
\ttfamily\small
\setlength{\parindent}{0pt}
\setlength{\parskip}{0pt}

% label width
\newcommand{\labw}{0.18\textwidth}
\newcommand{\txtw}{\dimexpr\textwidth-\labw\relax}

\noindent\makebox[\labw][l]{\promptlabel{System Prompt:}}%
\begin{minipage}[t]{\txtw}
You are a dialogue simulator acting as a user engaged in troubleshooting with an AI assistant.
\end{minipage}

\vspace{0.6ex}

\noindent\makebox[\labw][l]{\promptlabel{User Prompt:}}%
\begin{minipage}[t]{\txtw}
Your task is to generate the user’s reply at each turn, based on:
\begin{enumerate}[leftmargin=*, topsep=0pt, itemsep=0pt, parsep=0pt, partopsep=0pt]
\item User goal: What the user is trying to solve, achieve, or ask about.
\item Observable situation: troubles, conditions, status, or other observable facts described by the user.
\item Pre-established actions or checks: Actions, inspections, or operations that have already been completed before the conversation begins.
\item Reference dialogue: An example dialogue provided for guide language style and questioning behavior.
\end{enumerate}

\medskip
Simulation Rules:
\begin{enumerate}[leftmargin=*, topsep=0pt, itemsep=0pt, parsep=0pt, partopsep=0pt]
\item Respond only from the user's perspective.
\item Use the provided scenario context (goal, observable situation, and completed
   actions) when generating replies.
\item If the assistant asks about a condition the user can check or observe,
   respond based on the known scenario facts.
\item If the assistant asks about something the user does not know, ask for
   clarification or request guidance naturally.
\item If the question is new (not in the reference dialogue), infer a plausible reply based on the known context and reference dialogue style.
\item Keep the response concise and conversational, as a real user would.
\end{enumerate}

\medskip
% Prior information: \texttt{\{prio\_information\}}\\
% Reference dialogue: \texttt{\{ref\_dialog\}}\\
% Current conversation: \texttt{\{history\}}\\
% Unanswered questions: \texttt{\{unanswerd\_question\}}\\

\texttt{user\_goal: \{user\_goal\}\\
observable\_situation: \{observable\_situation\}\\
completed\_action\_or\_checks: \{completed\_actions\_or\_checks\}\\
}
AI Assistant: \texttt{\{input\}}\\
User simulator:
\end{minipage}

\end{promptbox}
\caption{The prompt template for user simulator.}
\label{fig:prompt_template_user_simulator}
\end{figure*}

\begin{figure*}[t]
\centering
\begin{promptbox}
\ttfamily\small
\setlength{\parindent}{0pt}
\setlength{\parskip}{0pt}

% Left label width (adjust if needed)
\newcommand{\labw}{0.18\textwidth}
\newcommand{\txtw}{\dimexpr\textwidth-\labw\relax}

\noindent\makebox[\labw][l]{\promptlabel{System Prompt:}}%
\begin{minipage}[t]{\txtw}
You are an assistant that determines whether the user answered a given yes/no question.
\end{minipage}

\vspace{0.6ex}

\noindent\makebox[\labw][l]{\promptlabel{User Prompt:}}%
\begin{minipage}[t]{\txtw}
Your job is to analyze the user's input and output one of the following:
\begin{itemize}\itemsep0pt
\item \texttt{yes} --- if the user clearly or implicitly answered \texttt{yes}
\item \texttt{no} --- if the user clearly or implicitly answered \texttt{no}
\item \texttt{trigger} --- Need to trigger a knowledge base to answer user's question. If the user did NOT answer the question and instead asked a new question or changed the topic.
\end{itemize}

\medskip
You MUST output only one word: \texttt{yes}, \texttt{no}, or \texttt{trigger}.

\medskip
You must carefully analyze what the question is asking; do not rely solely on the presence of positive or negative words in user input.

\medskip
Do not include any punctuation or extra words.

\medskip
Given question:\\
\texttt{\{question\}}

\medskip
User input:\\
\texttt{\{input\}}

\medskip
Your response:

\end{minipage}

\end{promptbox}
\caption{The prompt template for interactive flowchart reasoning in FLODIAL.}
\label{fig:prompt_template_flowchart_reasoning}
\end{figure*}

\begin{figure*}[t]
\centering
\begin{promptbox}
\ttfamily\small
\setlength{\parindent}{0pt}
\setlength{\parskip}{0pt}

% Left label width (adjust if needed)
\newcommand{\labw}{0.18\textwidth}
\newcommand{\txtw}{\dimexpr\textwidth-\labw\relax}

\noindent\makebox[\labw][l]{\promptlabel{System Prompt:}}%
\begin{minipage}[t]{\txtw}
You are a factual-consistency judge for dialogue generation.
Your job is to check whether each USER utterance in a predicted dialogue violates any known facts.
Known facts are ONLY the USER utterances from the given ground-truth dialogue.
Be conservative: only mark a violation when there is a clear contradiction with known facts.
If information is not mentioned in known facts, mark it as \texttt{"UNKNOWN"} rather than \texttt{"VIOLATION"}.
Return results in the required JSON format only.
\end{minipage}

\vspace{0.6ex}

\noindent\makebox[\labw][l]{\promptlabel{User Prompt:}}%
\begin{minipage}[t]{\txtw}
Given:
\begin{enumerate}[leftmargin=*, topsep=0pt, itemsep=0pt, parsep=0pt, partopsep=0pt]
\item \texttt{ground\_truth\_dialogue}: USER utterances inside it are the ONLY known facts.
\item \texttt{predicted\_dialogue}: evaluate EACH USER utterance for factual consistency against known facts.
\end{enumerate}

\medskip
Labels:
\begin{itemize}[leftmargin=*, topsep=0pt, itemsep=0pt, parsep=0pt, partopsep=0pt]
\item \texttt{CONSISTENT}: not contradicted by any known fact (may be a paraphrase or compatible detail)
\item \texttt{VIOLATION}: clearly contradicts at least one known fact
\item \texttt{UNKNOWN}: cannot be verified from known facts (neither supported nor contradicted)
\end{itemize}

\medskip
Rules:
\begin{itemize}[leftmargin=*, topsep=0pt, itemsep=0pt, parsep=0pt, partopsep=0pt]
\item Only judge USER utterances in \texttt{predicted\_dialogue}.
\item If a predicted user utterance contains multiple claims:
\begin{itemize}[leftmargin=*, topsep=0pt, itemsep=0pt, parsep=0pt, partopsep=0pt]
\item If ANY claim is \texttt{VIOLATION} $\Rightarrow$ utterance label is \texttt{VIOLATION}
\item Else if ANY claim is \texttt{UNKNOWN} $\Rightarrow$ utterance label is \texttt{UNKNOWN}
\item Else \texttt{CONSISTENT}
\end{itemize}
\end{itemize}

\medskip
\# Output (JSON only)
\begin{lstlisting}[basicstyle=\ttfamily\footnotesize,breaklines=true,columns=fullflexible,showstringspaces=false]
{
  "CONSISTENT": ["<pred_turn_id_1>", "<pred_turn_id_2>", "..."],
  "VIOLATION": ["..."],
  "UNKNOWN": ["..."]
}
\end{lstlisting}

\medskip

\medskip
\# ground\_truth\_dialogue\\
\texttt{\{GROUND\_TRUTH\_DIALOGUE\}}

\medskip
\# predicted\_dialogue\\
\texttt{\{PREDICTED\_DIALOGUE\}}
\end{minipage}

\end{promptbox}
\caption{The prompt of LLM judge for user simulator evaluation.}
\label{fig:factual_judge_prompt}
\end{figure*}

\label{sec:appendix}

% This is an appendix.

\end{document}